\documentclass[journal]{IEEEtran}

\ifCLASSOPTIONcompsoc
  \usepackage[nocompress]{cite}
\else
  \usepackage{cite}
\fi

\usepackage{times}
\usepackage{amsmath}
\usepackage{epsfig}
\usepackage{graphicx}
\usepackage{amsmath}
\usepackage{amssymb}
\usepackage{subfigure}
\usepackage{multirow}
\usepackage{algorithmicx}
\usepackage[table]{xcolor}
\usepackage{color, colortbl}
\usepackage[lined, ruled]{algorithm2e}
\usepackage[breaklinks=true,bookmarks=false]{hyperref}

\makeatletter
\DeclareRobustCommand\onedot{\futurelet\@let@token\@onedot}
\def\@onedot{\ifx\@let@token.\else.\null\fi\xspace}

\def\etal{\emph{et al}\onedot}

\newcommand{\fnz}[1]{\footnotesize{#1}}

\newcommand{\Cgray}[1]{\cellcolor[HTML]{D8D6D6}} 

\makeatother

\begin{document}
\title{Joint Person Re-identification and Camera Network Topology Inference in Multiple Cameras}

\author{Yeong-Jun~Cho, Su-A Kim*, Jae-Han Park*, Kyuewang Lee,~\IEEEmembership{Student Member,~IEEE}
        and~Kuk-Jin~Yoon,~\IEEEmembership{Member,~IEEE}
\IEEEcompsocitemizethanks{\IEEEcompsocthanksitem 
	This work was done when all authors were in Gwangju Institute of Science and Technology.
	* These authors contributed equally to this work.
	Yeong-Jun Cho, Jae-Han Park and Kuk-Jin Yoon are with the School of Electrical Engineering and Computer Science, Gwangju Institute of Science and Technology, South Korea. Currently, Su-A Kim is with Intel Visual Computing Institute Saarland Informatics Campus, Germany and Kyuewang Lee is with ASRI, Department of Electrical and Computer Engineering, Seoul National University. Kuk-Jin Yoon is a corresponding author. 
E-mail: kjyoon@gist.ac.kr}
}



\IEEEtitleabstractindextext{%
\begin{abstract}
	Person re-identification is the task of recognizing or identifying a person across multiple views in multi-camera networks.
	Although there has been much progress in person re-identification, person re-identification in large-scale multi-camera networks still remains a challenging task because of the large spatio-temporal uncertainty and high complexity due to a large number of cameras and people. To handle these difficulties, additional information such as camera network topology should be provided, which is also difficult to automatically estimate, unfortunately. 
	In this study, we propose a unified framework which jointly solves both person re-identification and camera network topology inference problems with minimal prior knowledge about the environments. 
	The proposed framework takes general multi-camera network environments into account and can be applied to online person re-identification in large-scale multi-camera networks.  
	In addition, to effectively show the superiority of the proposed framework, we provide a new person re-identification dataset with full annotations, named \texttt{SLP}, captured in the multi-camera network consisting of nine non-overlapping cameras.
	Experimental results using our person re-identification and public datasets show that the proposed methods are promising for both person re-identification and camera topology inference tasks.
\end{abstract}

}

\maketitle

\IEEEdisplaynontitleabstractindextext
\IEEEpeerreviewmaketitle

\ifCLASSOPTIONcompsoc

\fi

	\section{Introduction}
	
\begin{figure}[t]
		\centering
		\subfigure[Spatio-temporal uncertainties \newline between cameras]{\includegraphics[width=0.48\columnwidth]{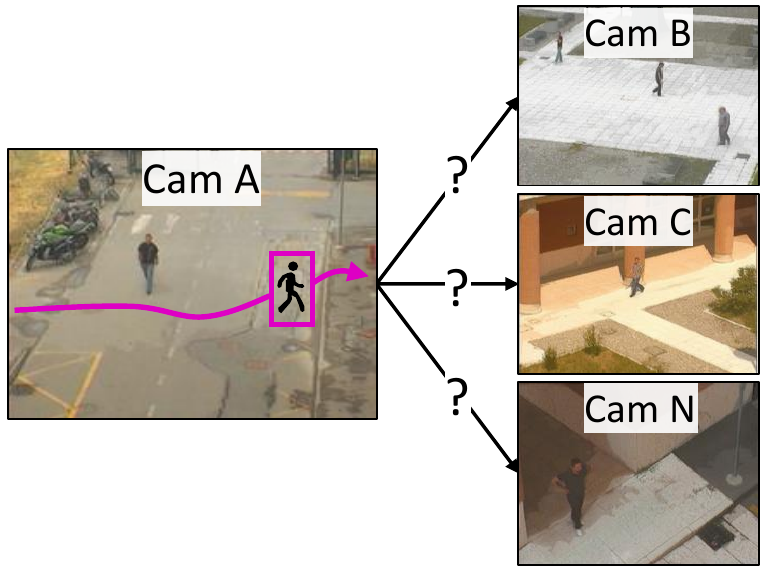}}
		\hspace{2pt}
		\subfigure[High complexity due to large numbers of cameras and people]{\includegraphics[width=0.48\columnwidth]{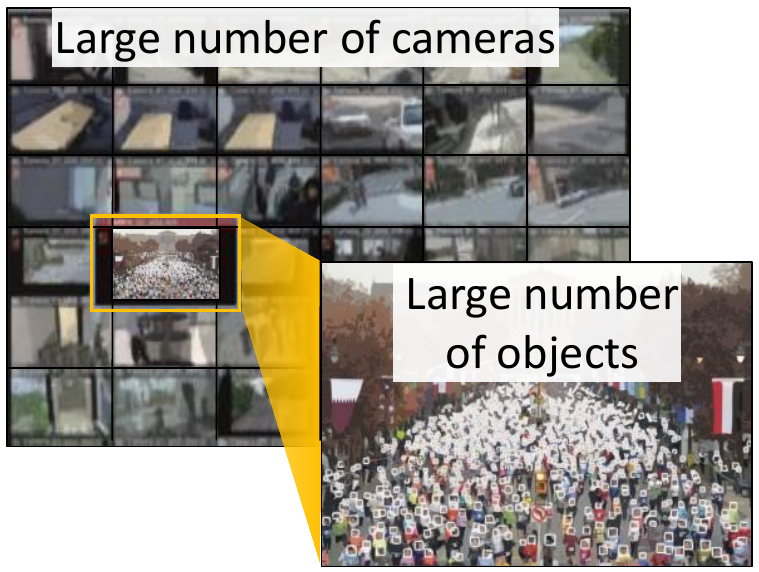}}
		\vspace{-0pt}
		\caption{Challenges of large-scale person re-identification}
		\label{fig_1}
		\vspace{-0pt}
	\end{figure}

	Nowadays, a large number of surveillance cameras installed at public places (e.g. offices, stations, airports, and streets) are used to prevent untoward incidents, monitor localities, or to track down evidences. In this context, multi-camera networks have been studied to monitor larger areas or specific individuals.
	Person re-identification, studied extensively over the past decade, is the task of \textit{automatically} recognizing and identifying a person using multiple views obtained from multi-camera networks. 
	Nevertheless, the re-identification in large-scale multi-camera networks still remains a challenging task because of the large spatio-temporal ambiguity and high complexity due to the large numbers of cameras and people. It becomes even more challenging when camera views do not overlap each other.
	As shown in Fig.~\ref{fig_1} (a), spatio-temporal uncertainty due to the unknown geometrical relationship between cameras in a multi-camera network makes re-identification difficult. Unless some prior knowledge about the camera network is given, re-identification should be done by thoroughly matching the person of interest with all the other people who appear in the camera network within the given time interval.
	This exhaustive search method is slow in general and generates unsatisfactory results in many cases because it is hard to find a correct match among a large number of candidates, as there might be many people having appearances similar to the person of interest.
	
	Most of the previous studies~\cite{farenzena2010person, davis2007information, ahmed2015improved} conducted an exhaustive search to re-identify people across multiple cameras, relying solely on the appearance of people.
	These methods work quite well when the numbers of cameras and people are small, but cannot effectively handle the aforementioned challenges in the large-scale problem illustrated in Fig.~\ref{fig_1} (b).
	To resolve the complexity problem and to improve the accuracy of the large-scale person re-identification, the number of matching candidates should be constrained and reduced by inferring and exploiting the spatio-temporal relation between cameras, referred to as the camera network topology.
	
	In recent years, several camera network topology inference methods~\cite{makris2004bridging, loy2010time} have been proposed for large-scale person re-identification.
	These methods infer the topology of a camera network based on the simple occurrence correlation between the people's entry and exit events.    
	However, no appearance-based validation is performed for topology inference. Hence, the inferred topology can be inaccurate in crowded scenes. Actually, it is a chicken-and-egg situation: the camera network topology can be more accurately inferred from the re-identification results while person re-identification can be improved by utilizing the camera network topology information.
	
	The main idea of this study is that the camera network topology inference and person re-identification can be achieved jointly while complementing each other. 
	Based on this idea, we propose a unified framework, which automatically solves both the problems together.
	To the best of our knowledge, this is the first attempt to solve both the problems jointly.
	In the proposed framework, we first infer the initial camera network topology using only highly reliable re-identification results obtained by the proposed multi-shot matching method. This initial topology is used to improve the person re-identification results, and the improved re-identification results are then used to refine the camera network topology. This procedure is repeated until the estimated camera network topology converges. Once we estimate the reliable camera network topology in the initialization stage, we utilize it for the online person re-identification and update the camera network topology with time. 
	
	To sum up, we propose a multi-shot person re-identification method that exploits a time-efficient random forest. (Sec.~\ref{subsec:RF_person_reid}).
	We also propose a fast and accurate camera network topology inference method in Sec.~\ref{subsec:topology_infer}.
	In Sec.~\ref{subsec:online_reid}, we describe a proposed online person re-identification method.
	The methods are effectively integrated in our framework. It is important to note that our proposed framework runs automatically with minimal prior knowledge about the environments.
	Besides the proposed methods, we also provide a new Synchronized Large-scale Person re-identification dataset named {\texttt{SLP}} (Sec.~\ref{sec:Pe-Lar_database}) captured in the multi-camera network consisting of nine non-overlapping cameras. 
	To validate our unified framework, we extensively evaluate the performance and compare it with state-of-the art methods.  
	
	Our previous study is \cite{cho2017unified}. Differences between the previous study and this study are as follows. 
	In this study, we propose and validate online person re-identification after the camera network topology initialization stage~(Sec.~\ref{subsec:online_reid}). During the online person re-identification, the camera network topology updates itself with time.
	We describe complete details of the proposed methods and provide more experimental results to explicitly validate our proposed methods. In Sec.~\ref{sec:exp}, we compare our proposed method with state-of-the-art methods in terms of both quantitative and qualitative performances, and show the results using another public dataset.
	
	\section{Previous Works}
	\label{sec:preivous}
	
	Person re-identification methods can be categorized into non-contextual and contextual methods as summarized in~\cite{bedagkar2014survey}.
	In general, non-contextual methods rely only on appearances of people and measure visual similarities between people to establish correspondences, while contextual methods exploit additional contexts such as human pose, camera parameters, geometry, and camera network topology.
	
	\subsection{Non-contextual Methods}
	In order to identify people across non-overlapping views, most studies rely on appearances of people by utilizing appearance-based matching methods with feature learning or metric learning techniques.	
	For feature learning, many studies~\cite{farenzena2010person, liu2012person,zhao2014learning} have tried to design visual descriptors to describe the appearance of people suitably.
	Regarding the metric learning, several methods such as KISSME~\cite{koestinger2012large} and LMNN-R~\cite{dikmen2011pedestrian} have been proposed and applied to the re-identification problem~\cite{davis2007information, weinberger2005distance, roth2014mahalanobis}. 
	Some studies \cite{koestinger2012large,roth2014mahalanobis} extensively evaluated and compared several metric learning methods and showed the effectiveness of the metric learning for re-identification.
	In order to utilize more plentiful appearance information, various multi-shot matching methods exploiting multiple appearances were proposed in~\cite{farenzena2010person, wang2014person, li2015multi}. In addition, person re-identification methods based on deep learning were proposed in~\cite{ahmed2015improved, yi2014deep}, which simultaneously learn both features and distance metrics.
	
	Although many non-contextual methods have improved the performance of person re-identification, the challenges such as spatio-temporal uncertainty between non-overlapping cameras and high computational complexity still remain.
	
	\subsection{Contextual Methods}
	Several studies~\cite{bak2015person,wu2015viewpoint, cho2016improving} using human pose priors have been proposed recently.
	These methods calibrate cameras to obtain camera parameters and estimate human poses. Person re-identification with estimated human poses can then be efficiently carried out. The methods mitigate ambiguities by seeking various poses of people.
	
	\begin{figure*}[t]
		\centering
		\vspace{-0pt}        
		\includegraphics[width=2.00\columnwidth]{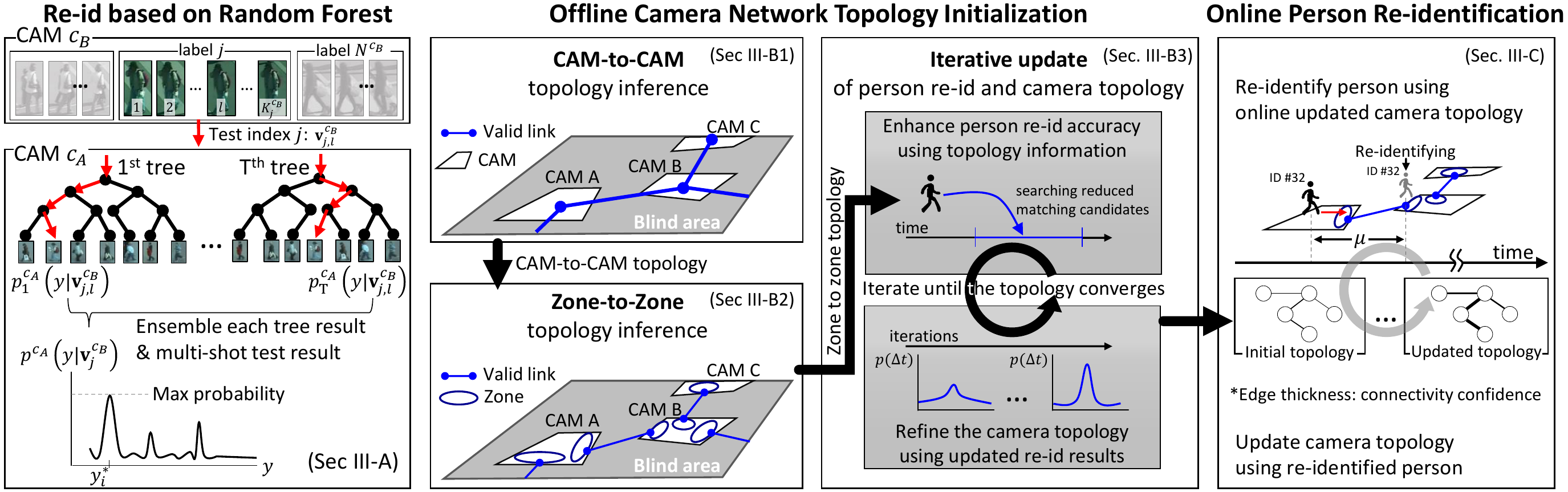} 		\vspace{-0pt}
		\caption{The proposed unified framework for person re-identification and camera network topology inference.}\vspace{-0pt}
		\label{fig_2}
	\end{figure*}
	
	To resolve the spatio-temporal ambiguities, many studies have tried to employ camera network topology and camera geometry. 
	Several studies~\cite{javed2003tracking,cai2014exploring,rahimi2004simultaneous} assume that the camera network topology information is available and show the effectiveness of the topological information.
	However, the topology is not available in the real-world scenario. Thus, many studies have tried to infer the camera network topology in an unsupervised manner.
	Makris~\textit{et al.}~\cite{makris2004bridging} proposed a topology inference method that simply observes the entry and exit events of the targets and measures the correlations between the events to establish the camera network topology. This method was applied and extended in~\cite{niu2006recovering,stauffer2005learning,chen2014object}. Similarly, Loy~\textit{et al.}~\cite{loy2010time,loy2012incremental} also proposed topology inference methods to understand multi-camera activity by measuring correlation or common information across simple activity patterns.
	
	The aforementioned topology inference methods~\cite{makris2004bridging, stauffer2005learning,chen2014object, loy2010time,loy2012incremental}, so-called event-based approaches are practical since they do not require any appearance matching steps such as re-identification or inter-camera tracking for topology inference.
	However, the topology inferred by the event-based approach can be inaccurate since the topology may be inferred from false event correlations. 
	Actually, false event correlation may easily occur when people pass through blind regions irregularly. 
	On the other hand, a few studies~\cite{cai2010recovering,martinel2016person} inferred the topology using person re-identification results. These methods tend to be more robust to noise than the event-based approaches. 
	However, they do not refine or update the initially inferred topology with time while performing re-identification. Thus, these methods may not fully exploit the topology information.

	\section{Proposed Unified Framework}
	\label{sec:proposed}
	Figure~\ref{fig_2} illustrates the proposed unified framework for person re-identification and camera network topology inference.
	In the proposed framework, we first train random forest-based person classifiers~(Sec.~\ref{subsec:RF_person_reid}) for efficient person re-identification. Subsequently, we jointly estimate and refine the camera network topology and person re-identification results~(Sec.~\ref{subsec:topology_infer}) using the trained random forests. Finally, we perform an online person re-identification using the inferred camera topology~(Sec.~\ref{subsec:online_reid}). 
	
	\subsection{Random Forest-based Person Re-identification}
	\label{subsec:RF_person_reid}
	Most of the previous studies mainly focused on enhancing the re-identification performance. However, when handling a large number of people, time complexity is also very important for building a practical re-identification framework. 
	To this end, we utilize a random forest algorithm~\cite{breiman2001random} for multi-shot person re-identification and incorporate it into our framework.
	We denote the $k$-th appearance of person $i$ in the camera $c_{A}$ as ${\mathbf{ v }}^{ c_{A} }_{i,k}$. A set of the appearances of people in the camera is expressed as,
	\vspace{-0pt}
	\begin{equation}
		\label{eq:1}
		\mathcal{D}^{c_{A}}={ \left\{ { \left( {\mathbf{ v }}^{ c_{A} }_{i,k},{ y }_{ i } \right)  } | { 1 \leq i \leq N^{c_{A}}, 1 \leq k \leq K^{c_{A}}_i}\right\}},
	\end{equation}
	where $y_{i}$ is the label of person $i$ (normally ${ y }_{ i }$ is set to $i$). $N^{c_{A}}$ is the number of people visible in camera $c_{A}$ and $K^{c_{A}}_i$ is the number of appearances of person $i$ in camera $c_{A}$. We then train a random forest classifier using the appearance set $\mathcal{D}^{c_{A}}$.
	After the random forest classifier is trained, we can estimate the probability distribution of classification by aggregating decision tree outputs as:
	$ p^{c_{A}}\left( { y }|{ \mathbf{v} } \right) =\frac { 1 }{ \mathcal{T} } \sum _{ \tau=1 }^{ \mathcal{T} }{ { p }^{c_{A}}_{ \tau }\left( y|\mathbf{v} \right)}, $
	where ${ p }^{c_{A}}_{ \tau }$, $\mathcal{T}$ denote a trained decision tree and the number of decision trees, respectively. 
	Testing a person $j$ (i.e. probe) in another camera ${c}_{B}$ to the trained random forest classifier is defined as,
	\vspace{-0pt}
	\begin{equation}
		\label{eq:3}
		p^{{c}_{A}}\left( { y }|{ {\mathbf{v}}^{{c}_{B}}_{ j,l } } \right) =\frac { 1 }{ \mathcal{T} } \sum _{ \tau=1 }^{ \mathcal{T} }{ { p }^{{c}_{A}}_{ \tau }\left( y|{\mathbf{v}}^{{c}_{B}}_{ j,l } \right)}. 
	\end{equation}

	\begin{figure}[t]
		\centering
		\includegraphics[width=1.00\columnwidth]{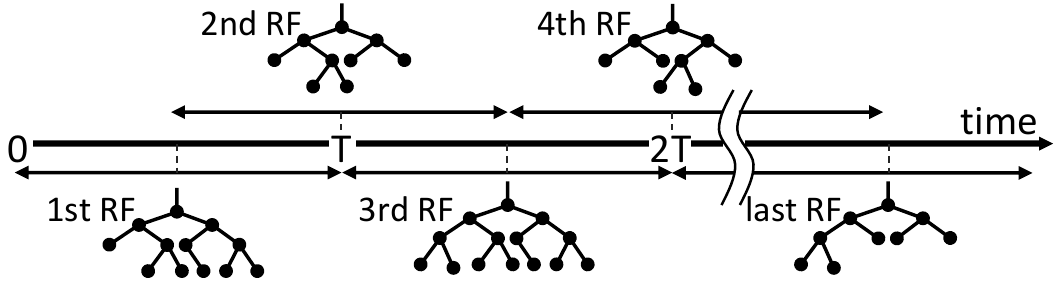}
		\vspace{-0pt}
		\caption{Training a series of random forest classifiers based on the overlapping time window $T$.}
		\vspace{-0pt}
		\label{fig_4}
	\end{figure}
	
	To obtain a multi-shot person re-identification result, we test multiple appearances of each person and average the multiple results as 
	\begin{equation}
	p^{{c}_{A}}( y |{ {\mathbf{v}}^{{c}_{B}}_{j} } ) =\frac { 1 }{ K_{j}^{c_{B}} }\sum _{ l=1 }^{ K_{j}^{c_{B}} }{ p^{{c}_{A}}( y |{ {\mathbf{v}}^{{c}_{B}}_{ j,l } } )  },
	\end{equation}
	where $K_{j}^{c_{B}}$ is the number of appearances of the probe.
	From the probability distribution $p^{{c}_{A}}\left( y |{ {\mathbf{v}}^{{c}_{B}}_{j} }\right)$, we choose a final matched label as:
	\begin{equation}
	{y}_{i}^{*}= \underset { y_{i}\in \left\{ 1,2,...,N^{c_{A}} \right\}}{\mathrm{ argmax }} p^{{c}_{A}}( y_{i} |{ {\mathbf{v}}^{{c}_{B}}_{j} } ).
	\end{equation}
	As the result of the multiple appearance matching test, we obtain a corresponding pair~($\mathbf{ v }_{{y}_{i}^{*}}^{{c}_{A}}$, $\mathbf{ v }_{j}^{ { c }_{ B } }$) between the camera $c_{A}$ and $c_{B}$. 
	Finally, we calculate a similarity score for the matching pair by selecting the smallest matching score as per the following equation~\cite{farenzena2010person}, 
	\begin{equation}
		S\left( \mathbf{ v }_{ {y}_{i}^{*} }^{ { c }_{ A } }, \mathbf{ v }_{ j }^{ { c }_{ B } } \right) = { e }^{ -\min _{ k,l }{ { \left\| \mathbf{ v }_{ y_{i}^{ * }, k }^{ { c }_{ A } }- \mathbf{ v }_{ j,l }^{ { c }_{ B } } \right\|  }_{ 2 } }}.
	\end{equation}
	The similarity score lies on $[0,1]$ and it is used in Sec.~\ref{subsec:topology_infer} for inferring the topology.

	\begin{figure*}[t]
		\centering
		\subfigure[CAM1 -- CAM2]{\includegraphics[width=0.8\columnwidth]{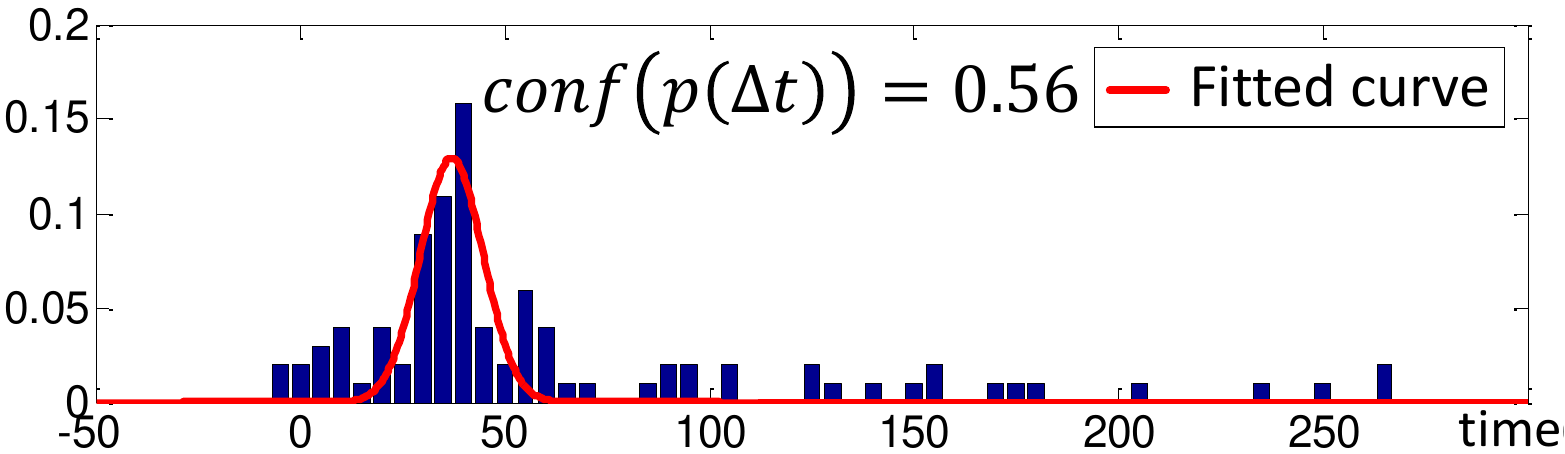}}
		\hspace{5pt}
		\subfigure[CAM1 -- CAM4]{\includegraphics[width=0.8\columnwidth]{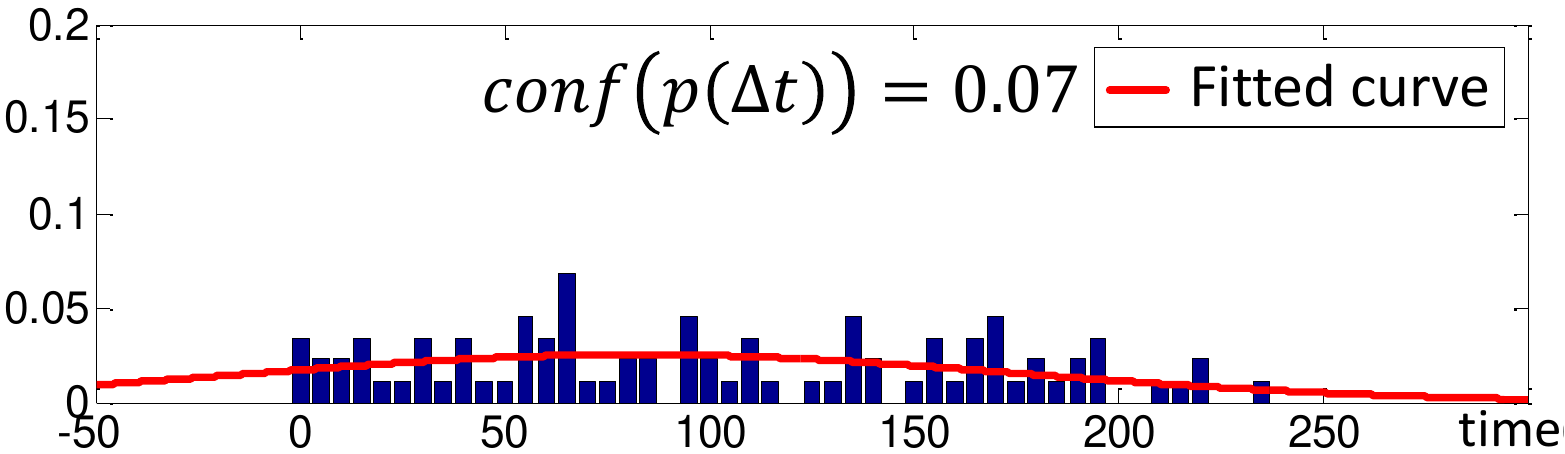}}
		\vspace{-0pt}
		\caption{Examples of estimated transition distributions with connectivity confidences}
		\vspace{-0pt}
		\label{fig_5}
	\end{figure*}

	The tree structure of the random forest method makes the multi-shot test very fast: its computational complexity is $O(NK\log { N })$\footnote{When analyzing the computational complexity, we assume that the number of objects in ${c}_{A}$ and ${c}_{B}$ are the same $(i,j=\left\{1,2,..,N\right\})$ and each object contains the same number of image patches $(k,l=\left\{1,2,..,K\right\})$. Details are described in APPENDIX.}. On the other hand, the conventional exhaustive re-identification approaches comparing every person pair involve higher computational cost: $O({N}^{2}{K}^{2})$.
	Over and above the superiority in terms of the computational cost, our method also delivers a high person re-identification accuracy as shown in Sec.~\ref{subsec:exp:cam_to_cam}.
	
	\subsection{Camera Network Topology Initialization}
	\label{subsec:topology_infer}
	
	Camera network topology represents spatio-temporal relations and connections between the cameras in the network.
	It involves the inter-camera transition distributions between two cameras, which denote the transition distributions of objects across cameras according to time, and represent the strength of connectivity between the cameras.
	In general, the topology is represented as a graph $G=\left(V, E\right)$, where the vertices $V$ denote the cameras and the edges $E$ denote the inter-camera transition distributions, as shown in Fig.~\ref{fig_10} (b).

	\subsubsection{CAM-to-CAM topology inference}
	\label{subsubsec:CAMCAM_con_analy}

	To build the CAM-to-CAM topology, we first estimate the transition distributions between the cameras. 
	In this study, we estimate the transition distributions based on the person re-identification results.
	Before performing person re-identification, we select key appearances among the multiple appearances of each person based on the method in~\cite{han2000new}.
	By selecting key appearances uncorrelated with each other, we can efficiently reduce the matching complexity and avoid over-fitting problems.
	
	After the key appearance selection, we split a whole group of people into several sub-groups according to their timestamps and train a series of random forest classifiers with overlapping time window $T$ as shown in Fig.~\ref{fig_4}.
	Next, we search correspondences of people who disappeared in a camera using the trained random forest classifiers of other cameras. 
	Initially, we have no transition distributions between cameras to utilize. 
	Hence, we consider every pair of cameras in the camera network and search for correspondences within a wide time interval. 
	When a person disappears at time $t$ in a certain camera, we search the correspondence of the person from the other cameras within time range $[t-T,t+T]$. 
	If there are multiple random forest classifiers overlapping with the time range, we test all random forest classifiers and select the most reliable one among them as a correspondence. Thanks to our overlapping search scheme, we can alleviate the risk of failure in correspondence search.
	
	When these initial correspondences are given, we only use highly reliable correspondences to infer transition distributions between cameras.
	We regard a correspondence as a reliable one with a high similarity score when $S( \mathbf{ v }_{ { y }_{i}^{ * } }^{ { c }_{ A } }, \mathbf{ v }_{ j }^{ { c }_{ B } })> \theta_{sim}$.
	The inference procedure of transition distribution is as follows:
	(1) Calculate the time difference between correspondences and make a histogram of the time difference. 
	(2) Normalize the histogram using the total number of reliable correspondences. 
	We denote the transition distribution as $p\left(\Delta t\right)$.
	Figure~\ref{fig_5} shows two distributions: Fig.~\ref{fig_5}~(a) is obtained from a pair of cameras having a strong connection, and Fig.~\ref{fig_5}~(b) from a pair of cameras having a weak or no connection.

	\noindent \textbf{Connectivity Check} \quad Based on the estimated transition distributions, we automatically identify whether a pair of cameras is connected or not. We assume that the transition distribution follows a normal distribution if there is a topological connection.
	Based on this assumption, we fit a Gaussian model $N(\mu,\sigma^2)$ to the distribution $p\left(\Delta t\right)$ and measure the connectivity of a pair of cameras based on the following observations:

	\noindent $\bullet$ \textbf{Variance of $p\left(\Delta t\right)$}: In general, most of the people reappear after a certain transition time $\mu$; therefore, the variance in the transition distribution $(\sigma^2)$ is not extremely large and the distribution shows a clear peak.  
	
	\noindent $\bullet$ \textbf{Fitting error}: Even though the distribution is obtained from a pair of cameras with a weak connection, the variance in the distribution can be small and the distribution can have a clear peak due to noise. 
	In order to measure the connectivity robust to noise, we consider the model fitting error $\mathcal{E}\left( p\left(\Delta t\right) \right)\in[0,1]$ calculated by R-squared statistics.   
	
	Based on the above observations, we newly define a connectivity confidence between a pair of cameras as
	\begin{equation}
		conf\left( p\left(\Delta t\right) \right) = e^{-\sigma}\cdot \left(1-\mathcal{E}\left( p\left(\Delta t\right) \right)\right).
	\end{equation}
	The connectivity confidence lies on $[0, 1]$. 
	We regard a pair of cameras as a valid link when $conf\left(p\left(\Delta t\right)\right)>\theta_{conf}$.
	Compared to a previous method~\cite{makris2004bridging}, which only takes the variance of a distribution into account, our method is more robust to noise of distributions.
	Using the defined confidences, we check every pair of cameras and reject invalid links as shown in Fig.~\ref{fig_10}. We can see that many camera pairs in the camera network have weak connections. Therefore, we can greatly reduce computation time and save resources. Only the valid pairs of cameras proceed to the next step.

	Through this topology inference procedure, CAM-to-CAM topology is represented as
	\begin{equation}
	\begin{split}
		G_{cam} & =\left(V_{cam}, E_{cam}\right), \\
		V_{cam} & \in \left\{{c_i}|{ 1\le i\le N_{cam}  } \right\} , \\
		E_{cam} & \in \left\{{p^{ij}\left(\Delta t\right)}|{ 1\le i\le N_{cam}, 1\le j\le N_{cam}, i\neq  j} \right\}, 
	\end{split}
	\end{equation}
	where $N_{cam}$ is the number of cameras in the camera network, $c_i$ is a i$^{th}$ camera, and $p^{ij}\left(\Delta t\right)$ is a transition distribution between $c_i$ and $c_j$.

	\vspace{-0pt}

	\begin{figure*}[t]
		\centering
		\subfigure[Iteration 1]{\includegraphics[width=0.53\columnwidth]{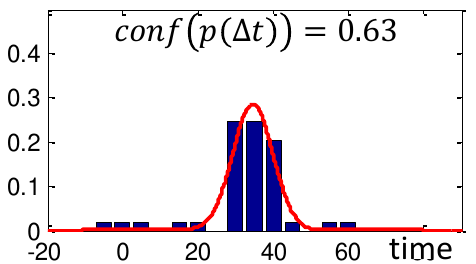}}
		\hspace{3pt}
		\subfigure[Iteration 2]{\includegraphics[width=0.53\columnwidth]{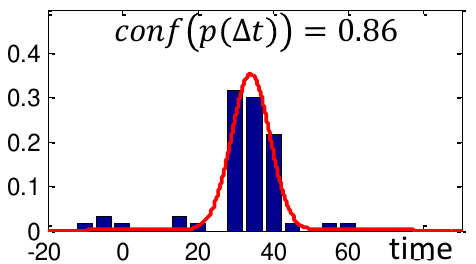}}
		\hspace{3pt}
		\subfigure[Iteration 5]{\includegraphics[width=0.53\columnwidth]{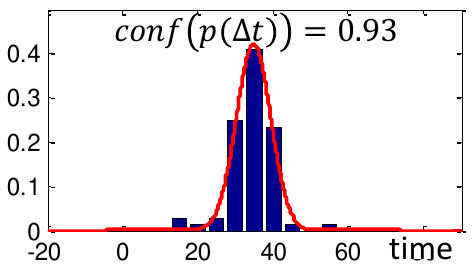}}
		\vspace{-0pt}
		\caption{Iterative update of the topology and re-identification results. Re-id accuracy is improved from 23.5\% to 66.31\%.}
		\vspace{-0pt}
		\label{fig_7}
	\end{figure*}

	\vspace{0pt}			
	\subsubsection{Zone-to-Zone topology inference} \label{subsubsec:ZonZon_con_analy}
	
	In this step, we estimate transition distributions between zones in cameras and build a Zone-to-Zone topology.
	For each camera, a set of entry and exit zones is automatically learned by \cite{makris2003automatic}.
	Note that we only consider exit-to-entry zone pairs when two zones belong to different cameras. 
	Other pairs of zones such as exit-to-exit, entry-to-entry, and entry-to-exit are not considered.
	
	A person who disappeared at an exit zone at time $t$ is likely to appear at an entry zone in a different camera within a certain time interval $T$. Therefore, we search the correspondence of the disappeared person from the entry zones in different cameras within the time range $[t,t+T]$.
	Similarly, as in Sec.~\ref{subsubsec:CAMCAM_con_analy}, we train a series of random forest classifiers for each entry zone and measure connectivity confidences of all possible pairs of zones using only reliable correspondences.
	Through this step, many invalid pairs of zones between the cameras can be ignored, and we can formulate Zone-to-Zone topology as:
	\begin{equation}
	\begin{split}
	G_{zone} = &\left(V_{zone}, E_{zone}\right), \\
	V_{zone} \in & \left\{{c_{i(k)}}|{ 1\le i\le N_{cam}  }, 1 \le k \le Z_{i} \right\} , \\
	E_{zone} \in & \{{p^{ij}_{(kl)}\left(\Delta t\right)}|{ 1\le i\le N_{cam}, 1\le j\le N_{cam}, } \\
	             & \qquad\qquad\quad  i\neq j, 1\le k \le Z_i, 1\le l \le Z_j \}	
	\end{split}
	\end{equation}
	where $N_{cam}$ is the number of cameras in the camera network, $Z_i$ is the number of zones of i$^{th}$ camera.
	$c_{i(k)}$ is a k$^{th}$ zone of the i$^{th}$ camera.
	$p^{ij}_{(kl)}\left(\Delta t\right)$ is a transition distribution between $c_{i(k)}$ and $c_{j(l)}$.
	In the next section, we iteratively update the valid links between zones and build a camera topology map of the camera network.

	\subsubsection{Iterative update of person re-identification \\ and camera network topology} 
	\label{subsubsec:iter_topology_infer}
	
	After conducting the Zone-to-Zone topology inference, we obtain an initial topology map between every pair of zones in the camera network.
	However, the initial topology map can be inaccurate, because it is inferred from noisy initial re-identification results.
	Although we consider only highly reliable re-identification results to infer the initial topology, the re-identification results are erroneous because we cannot exploit the topology information initially for re-identification.
	
	As mentioned above, camera network topology information and re-identification results can complement each other. The inferred topology of the camera network can enhance the person re-identification performance, and the person re-identification can assist the topology inference. Therefore, we update person re-identification results and the camera network topology in an iterative manner.
	Assuming that we have a transition distribution $p\left(\Delta t\right) \sim  N(\mu,\sigma^2)$ between two zones. Each zone belongs to different cameras. Then, the iterative update procedure is as follows:	
	
	\begin{algorithm}[t]
		\KwData{Image sequences and object identities in each view during initialization stage}
		\KwResult{Re-identification result (people correspondences), Camera network topology\newline}

		\For{every camera pair}{
			
			$G_{cam}$ = CAM-to-CAM topology inference;\newline
						
			\If{$conf\left(p^{ij}\left(\Delta t\right)\right)>\theta_{conf}$}{
				$G_{zone}$ = Zone-to-Zone topology inference;\newline
				
				\If{$conf\left(p^{ij}_{(kl)}\left(\Delta t\right)\right)>\theta_{conf}$}{
					\While{$p^{ij}_{(kl)}\left(\Delta t\right)$ converging}{
						
						$T^{ij}_{(kl)} = \frac { 1 }{ 1-\mathcal{E}\left( p^{ij}_{(kl)}\left(\Delta t\right) \right) } \left(T^{ij}_{(kl)U}-T^{ij}_{(kl)L} \right)$\\
						
						$G_{zone}$ = Topology update;
						
					}
				}
			}				
		}
		
		\caption{Camera network topology initialization}
		\label{alg1}
	\end{algorithm}
	
	\begin{itemize}[]
		\item \textbf{Step 1.} Update the time window $T$. 
		The initial time window $T$ was set quite wide, but now we can narrow it down based on the inferred topology. 
		First, we find the lower and upper time bounds $\left(T_{L},T_{U}\right)$ of the transition distribution $p\left(\Delta t\right)$ with a constant $R$ as,
		\begin{equation}
			p\left( T_{L}\le \Delta t\le T_{U} \right) = { \frac{R}{100} }.
		\end{equation}
		We set $R$ as $95$ following 3-sigma rule in order to cover a majority of the distribution (95\%) and ignore some outliers (5\%). Then, using the time bounds obtained, the time window $T$ is updated as,
		\begin{equation}
			\label{eq:6}
			T = \frac { 1 }{ 1-\mathcal{E}\left( p\left(\Delta t\right) \right) } \left(T_{U}-T_{L} \right),
		\end{equation}
		where $\mathcal{E}\left( p\left(\Delta t\right) \right)$ is a Gaussian model fitting error rate. When the fitting error is high, the time window $T$ becomes large. Thanks to our update strategy, we can avoid the over-fitting of the topology during the update steps.
		\item \textbf{Step 2.} Re-train a series of random forests of an entry zone with the updated time window $T$ as shown in the Fig.~\ref{fig_4}. 
		\item \textbf{Step 3.} Find correspondences of the disappeared people at an exit zone. Based on the topology, a person who disappeared at time $t$ at the exit zone is expected to appear around the time $(t+\mu)$ at the entry zone of the other camera. 
		Using the topological information, we search the correspondence of the person from a trained random forest having the center of time slot close to $(t+\mu)$.
		\item \textbf{Step 4.} Update the transition distribution using reliable correspondences with a high similarity score $S( \mathbf{ v }_{ { y }_{i}^{ * } }^{ { c }_{ A } }, \mathbf{ v }_{ j }^{ { c }_{ B } } )>\theta_{sim}$.
	\end{itemize}
		
	This procedure~(Step 1 -- Step 4) is repeated until the transition distribution converges. 
	We perform the procedure for all Zone-to-Zone pairs in the camera network topology $G_{zone}$.
	Consequently, the camera topology map in the camera network is refined. If the topology does not change anymore or shows subtle changes during several iterations, we stop the procedure. The above procedure improves the performance of re-identification as well as the accuracy of the topology inference, as shown in Fig.~\ref{fig_7}.
	The overall algorithm of the proposed camera network topology initialization process is summarized in Algorithm.~\ref{alg1}.

	\begin{algorithm}[t]
		\KwData{Image sequences, Object identities in each view, \\ Camera network topology}
		\KwResult{Re-identification result (people correspondences), Camera network topology \newline}
		
		\For{every zone pair}{
		
			\If{$conf\left(p^{ij}_{(kl)}\left(\Delta t\right)\right)>\theta_{conf}$}{

				Find correspondences;
				
				$p'^{ij}_{(kl)}(\Delta t)$ = Transition distribution update; \newline

				\If{$\sum _{ \Delta t }^{  }{ \left( \left| p^{ij}_{(kl)}(\Delta t) - p'^{ij}_{(kl)}(\Delta t)  \right|  \right)}>0.1$}{
				{\small \% Fit Gaussian model to $p'^{ij}_{(kl)}(\Delta t)$}
				$p^{ij}_{(kl)}(\Delta t) = p'^{ij}_{(kl)}(\Delta t)\sim N(\mu,\sigma^2)$;
				}
			
			}				
		}	
		
		\caption{Online person re-identification and \newline camera network topology update}
		\label{alg2}
	\end{algorithm}

	\begin{table}[t]
		\centering
		\caption{List of public person re-identification datasets. We referred to the list of person re-identification datasets in~\cite{Karanam2016peid_review}.} \vspace{-0pt}
		\label{tab_2}
		{
			\setlength\tabcolsep{2.0pt}  
			\begin{tabular}{r|r|r|r|c|c|c|c}
				\noalign{\hrule height 1pt}
				\multicolumn{1}{c|}{dataset}           & \multicolumn{1}{c|}{\textit{year}}        & \multicolumn{1}{c|}{\# \textit{IDs}}                     & \multicolumn{1}{c|}{\# \textit{cam}}              & \multicolumn{1}{c|}{\begin{tabular}[c]{@{}c@{}}\textit{multi}\\\textit{-shot}\end{tabular}} & \multicolumn{1}{c|}{\begin{tabular}[c]{@{}c@{}}\textit{back}\\\textit{ground}\end{tabular}} & \multicolumn{1}{c|}{\textit{traj.}}     & \multicolumn{1}{c}{\textit{sync}} \\ \hline
				\texttt{VIPeR}\cite{gray2007evaluating}             & 2007    & 632     & 2    &            &            &            &            \\ \hline
				\texttt{QMUL iLIDS}\cite{Zheng2009associating}      & 2009    & 119     & 2    & \checkmark &            &            &            \\ \hline
				\texttt{GRID}\cite{loy2010time}                     & 2009    & 250     & 8    &            &            &            &            \\ \hline
				\texttt{CAVIAR4ReID}\cite{cheng2011custom}          & 2011    & 72      & 2    & \checkmark &            &            &            \\ \hline
				\texttt{3DPeS}\cite{baltieri2011_308}               & 2011    & 192     & 8    & \checkmark & \checkmark &            &            \\ \hline
				\texttt{PRID2011}\cite{hirzer11a}                   & 2011    & 934     & 2    & \checkmark & \checkmark & \checkmark &            \\ \hline
				\texttt{V47}\cite{wang2011re}                       & 2011    & 47      & 2    & \checkmark &            &            &            \\ \hline
				\texttt{WARD}\cite{Martinel2012Re}                  & 2012    & 70      & 4    & \checkmark & \checkmark & \checkmark &            \\ \hline
				\texttt{Softbio}\cite{bialkowski2012database}       & 2012    & 150     & 8    & \checkmark & \checkmark & \checkmark &            \\ \hline
				\texttt{CUHK 2}\cite{li2013locally         }        & 2013    & 1816    & 2    &\checkmark  &            &            &            \\ \hline
				\texttt{CUHK 3}\cite{li2014deepreid        }        & 2014    & 1360    & 2    &\checkmark  &            &            &            \\ \hline
				\multirow{2}{*}{\texttt{NLPR MCT}\cite{Chen2015An}} & \multirow{2}{*}{2014}    & 235	    & 3          &\multirow{2}{*}{\checkmark}&\multirow{2}{*}{\checkmark}&\multirow{2}{*}{\checkmark}&\multirow{2}{*}{\checkmark} \\ \cline{3-4}
				&         & 255	    & 3    &            &            &            &            \\ \hline
				\texttt{RAiD}\cite{Das2014}                         & 2014    & 43      & 4    &\checkmark  &            &            &            \\ \hline
				\texttt{iLIDS-VID}\cite{wang2014person}             & 2014    & 300     & 2    &\checkmark  &            & \checkmark &            \\ \hline
				\texttt{Market1501}\cite{zheng2015scalable}         & 2015    & 1501    & 6    &\checkmark  &            &            &            \\ \hline
				\texttt{PRW}\cite{zheng2016person}                  & 2016    & 932     & 6    &\checkmark  & \checkmark &            &            \\ \hline  
				\texttt{MARS}\cite{zheng2016mars}                   & 2016    & 1261    & 6    &\checkmark  &            & \checkmark &            \\ \hline
				\texttt{DukeMTMC4ReID}\cite{gou2017dukemtmc4reid,ristani2016MTMC}& 2017    & 1852    & 8    &\checkmark  & 			  &            & \checkmark \\ \hline						
				\texttt{Airport}\cite{Karanam2016peid_review}       & 2017    & 9651    & 6    &\checkmark  &            &            &            \\ \hline \hline
				\textbf{\texttt{SLP}}                               & 2017    & 2632    & 9    &\checkmark  & \checkmark & \checkmark & \checkmark \\ \noalign{\hrule height 1pt}
			\end{tabular}}
			\vspace{-0pt}
	\end{table}

	\subsection{Online Person Re-identification \\ and Camera Network Topology Update}
	\label{subsec:online_reid}
	
	Once we estimate the reliable camera network topology in the initialization stage, we utilize it for the online person re-identification and online camera network topology update.
	The proposed online process is as follows: 
	
		\begin{table*}[t]
			\centering
			\caption{Details of our new dataset: \texttt{SLP}.} \vspace{-0pt}
			\label{tab_1}
			{
				\setlength\tabcolsep{3.2pt}  
				\begin{tabular}{r||c|c|c|c|c|c|c|c|c|c}
					\noalign{\hrule height 1pt}
					Index        & CAM 1         & CAM 2           & CAM 3            & CAM 4           & CAM 5           & CAM 6           & CAM 7          & CAM 8        & CAM 9    & Total \\ \hline\hline
					\# ID        &  256          & 661             & 1,175            & 243             & 817             & 324             & 516            & 711          & 641      &  2,632 \\ 
					\# frames    &  19,545       & 65,518          & 104,639          & 41,824          & 78,917          & 79,974          & 93,978         & 53,621       & 42,347   &  580,363 \\ 
					\# annotated box & 47,870    & 205,003         & 310,262          & 65,732          & 307,156         & 160,367         & 78,259         & 176,406      & 117,087  &  1,468,142 \\ 
					Resolution   &\fnz{720$\times$480}&\fnz{1920$\times$1080}&\fnz{1920$\times$1080}&\fnz{1440$\times$1080}&\fnz{1440$\times$1080}&\fnz{1440$\times$1080}&\fnz{1920$\times$1080}  &\fnz{1920$\times$1080}&\fnz{1920$\times$1080}  & -- \\ 
					Duration     & 2h 13m        & 2h 12m              & 2h 22m              & 2h              & 2h 21m              & 2h              & 2h 38m              & 2h 29m             & 2h 28m     &  --  \\ \noalign{\hrule height 1pt} 
				\end{tabular}
			}
			\vspace{-0pt}
		\end{table*}

		\begin{figure*}[t]
			\centering
			\subfigure[Layout of a camera network]{\includegraphics[height=0.42\columnwidth]{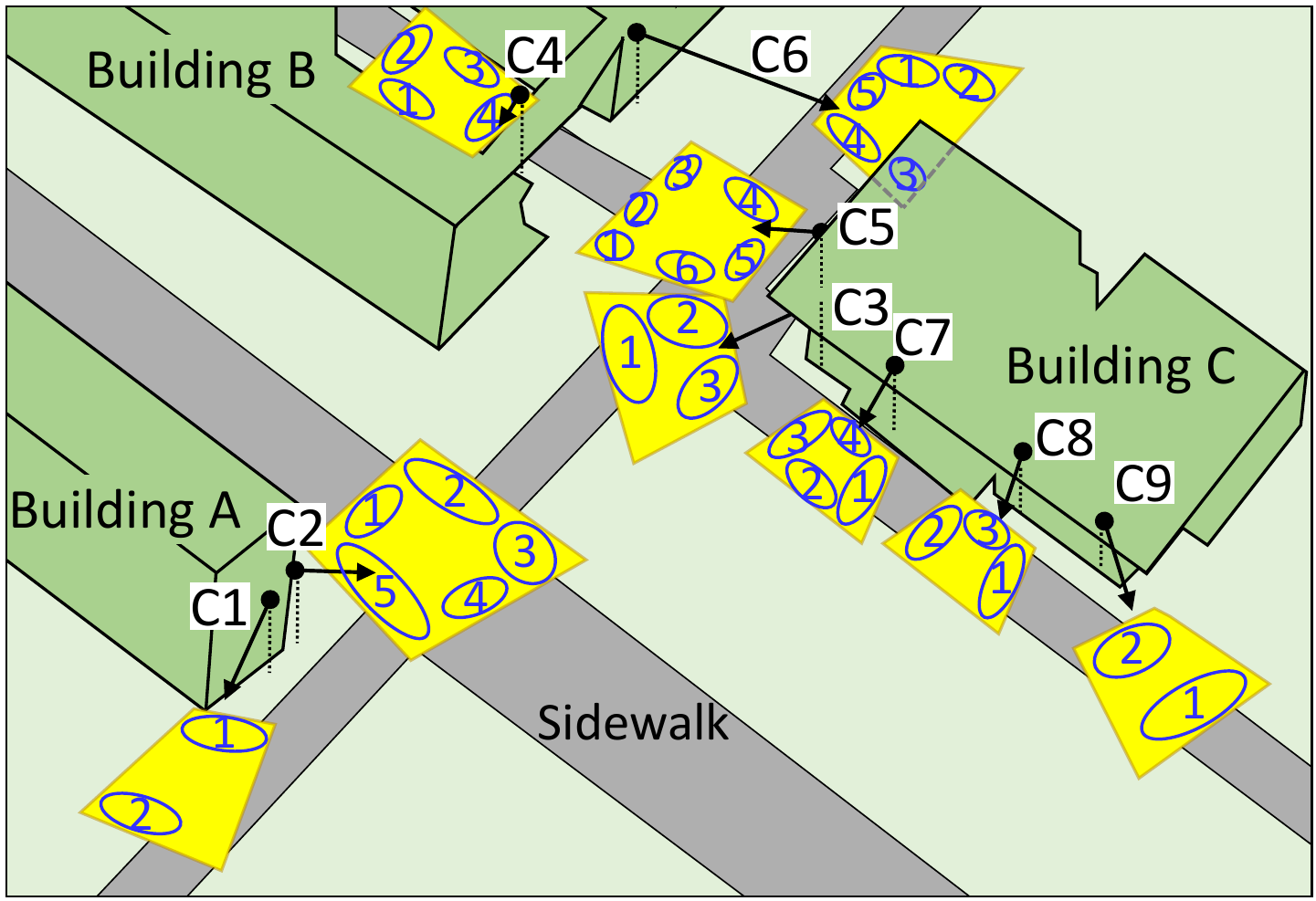}}
			\subfigure[Example frames of nine cameras~(CAM1 -- CAM9)]{\includegraphics[height=0.42\columnwidth]{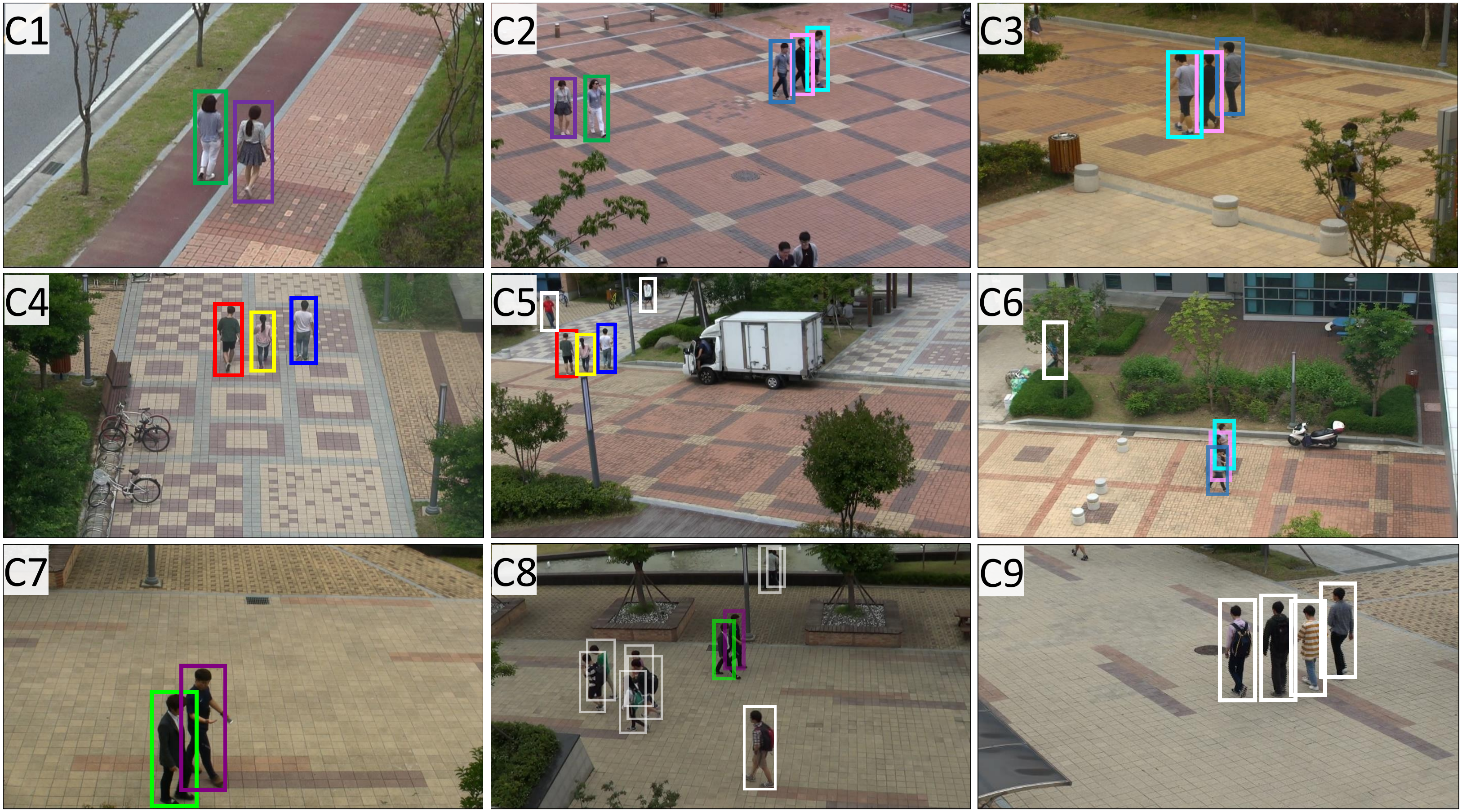}}
			\subfigure[Numbers of true matching pairs]{{\scriptsize
					\setlength\tabcolsep{1.2pt}  
					\renewcommand{\arraystretch}{1.22} 
					\begin{tabular}[b]{c||c|c|c|c|c|c|c|c|c}
						\noalign{\hrule height 1pt}
						CAM    & \textit{1}   &  \textit{2}   & \textit{3}   & \textit{4}   & \textit{5}   & \textit{6}   & \textit{7}    & \textit{8}   & \textit{9}   \\ \hline\hline
						\textit{1}    &\cellcolor[HTML]{C0C0C0}256   &  227  & 206   &  0   &  23   & 28   & 19   & 1  & 1\\ \hline
						\textit{2}    & 227 & \cellcolor[HTML]{C0C0C0}661   &  571  &  2  &  82  &  84   &  43  &  1 & 1\\ \hline
						\textit{3}    & 206 & 571 & \cellcolor[HTML]{C0C0C0}1175  &  98  & 568   &  90  &  168  &  57 & 56\\ \hline
						\textit{4}    & 0   & 2   & 98  & \cellcolor[HTML]{C0C0C0}243   &  155  &  0  &  37  &  16 & 16 \\ \hline
						\textit{5}    & 23  & 82  & 568 & 155 & \cellcolor[HTML]{C0C0C0}817   &  61  &  109  & 55 & 54\\ \hline
						\textit{6}    & 28  & 84  & 90  & 0   & 61  & \cellcolor[HTML]{C0C0C0}324   &  0  & 0  & 0\\ \hline
						\textit{7}    & 19  & 43  & 168 & 37  & 109 & 0  & \cellcolor[HTML]{C0C0C0}516   &  281 & 269 \\ \hline
						\textit{8}    & 1   & 1   & 57  & 16  & 55  & 0  & 281 & \cellcolor[HTML]{C0C0C0}711   & 633 \\ \hline
						\textit{9}   & 1   & 1   & 56  & 16  & 54  & 0  & 269 & 633 & \cellcolor[HTML]{C0C0C0}641 \\ 
						\noalign{\hrule height 1pt}
					\end{tabular}} } \vspace{-0pt}
					\caption{A new synchronized large-scale person re-identification dataset: \texttt{SLP}.}
					\label{fig_8} \vspace{-0pt}
				\end{figure*}

	\begin{itemize}[]
		\item \textbf{Step 1.} Pick out matching candidates of a person based on the initialized camera network topology. 
		When a person disappeared at a time $t$, a set of people appearing at a different camera within $t + T_L\le t+\mu \le t + T_U$ are the matching candidates. 
		The lower and upper time bounds ($T_L, T_U$) are evaluated using the method described in Sec.~\ref{subsubsec:iter_topology_infer}.
		\item \textbf{Step 2.} Find a correspondence among the matching candidates. For efficiency, we only train random forest classifiers where the number of matching candidates is large unless we find the correspondence based on exhaustive search.
		\item \textbf{Step 3.} Update the transition distribution. 
		We use the matching result with a high similarity score ($>\theta_{sim}$) to update the transition distribution, adding the new matching result to $p(\Delta t)$ and normalizing the distribution. The updated transition distribution is represented as $p'(\Delta t)$.
		\item \textbf{Step 4.} Fit a new Gaussian model to the updated transition distribution $p'(\Delta t)$. The cost of model fitting is quite high. 
		Hence, we do not always fit the Gaussian model, but only when the difference between the initial distribution $p\left( \Delta t \right)$ and the updated distribution $p'\left( \Delta t \right)$ is sufficiently large.
		To this end, we measure the distribution difference as $\sum _{ \Delta t }^{  }{ \left( \left| p\left( \Delta t \right) -p'\left( \Delta t \right)  \right|  \right)}$ and fit the model when the value is larger than 0.1 (i.e., 10 \% variation). Unless the value is larger than the threshold, $p'\left( \Delta t \right)$ merely cumulates the update.
	\end{itemize}
	We perform the aforementioned procedure for all Zone-to-Zone pairs in $G_{zone}$. The overall algorithm of the proposed online process is summarized in Algorithm.~\ref{alg2}.

	The online topology update step is necessary and useful for the following reasons: First, the pose of a camera can change due to vibration, heavy wind, and so on. Even a slight change in pose can cause a considerable viewpoint change, and it makes the initialized topology inaccurate. Second, people transition time between cameras can change temporarily due to many factors such as sidewalk conditions, weather, and unusual movement of crowd (e.g. parade, march). The online topology update can effectively handle the above issues. 
	
	Thanks to the accurate topology initialization~(Sec.~\ref{subsec:topology_infer}) and the online procedure~(Sec.~\ref{subsec:online_reid}), the camera network topology can be maintained automatically with time.
	When a new camera is added to an existing camera network during the online stage, the camera network topology initialization operates in parallel and the topology is updated efficiently.

	\section{New Person Re-identification Dataset: \texttt{SLP}}	
	\label{sec:Pe-Lar_database}

	To validate the performance of person re-identification methods, numerous datasets have been published. Table.~\ref{tab_2} shows a list of the public person re-identification datasets with their various characteristics, which have been constructed for specific scenarios. For example, {\texttt{VIPeR}}\cite{gray2007evaluating} was constructed by two cameras and contains 632 people. Each camera provides one single image for one person. 
	On the contrary, {\texttt{Softbio}}~\cite{bialkowski2012database} includes 150 people captured from eight cameras. In this dataset, each camera provides multiple images for a person~(\textit{multi-shot}).
	However, despite the availability of numerous published datasets, none of them reflect practical large-scale surveillance scenarios, 	in which (1) multiple observations (\textit{multi-shot}) and moving paths (\textit{traj.}) for each person from each camera are available, and (2) both the number of people~(\textit{\# IDs}) and the cameras~(\textit{\# cam}) is large.

	Most of the public datasets include a small number of people~(\textit{\# IDs} $<200$)~\cite{Zheng2009associating, cheng2011custom, baltieri2011_308, wang2011re, Martinel2012Re, bialkowski2012database, Das2014} or cameras~(\textit{\# cam} $<5$)~\cite{gray2007evaluating, Zheng2009associating, cheng2011custom, hirzer11a, wang2011re, li2013locally,li2014deepreid, Chen2015An,Das2014, wang2014person} as shown in Table.~\ref{tab_2}. Some datasets provide just a single shot of each person~\cite{gray2007evaluating, loy2010time}. Moreover, many datasets do not provide the required annotation with information about people (\textit{traj.}) such as their movement paths~\cite{Zheng2009associating, cheng2011custom, baltieri2011_308, wang2011re, li2013locally,li2014deepreid, zheng2015scalable, zheng2016person ,Karanam2016peid_review} but only give cropped multi-shot appearances. Furthermore, there are only two datasets that provide camera synchronization information or time stamps of all frames~(\textit{sync})~\cite{Chen2015An, gou2017dukemtmc4reid}. Recently, M. Gou~\etal~\cite{gou2017dukemtmc4reid} provided an re-identification dataset named \texttt{DukeMTMC4ReID}. The dataset covers real large-scale surveillance scenarios (e.g., synchronized cameras, large number of cameras and people). However, it does not give the trajectory of each person but only gives cropped multi-shot images. If the dataset provides additional information such as \textit{background} and \textit{traj.}, it will be very helpful for other researchers and can be fully exploited.
	
	In this study, we provide a new Synchronized Large-scale Person re-identification dataset called \texttt{SLP} constructed for practical large-scale surveillance scenarios. The main characteristics of our dataset are as follows:

	\renewcommand\labelitemi{\small$\bullet$}
	\begin{itemize}[]
		\item \textbf{The number of people} (\# \textit{IDs}):
		The total number of people in the dataset is 2,632. Among them, 1,737 people appear in at least two cameras.
		The details of the matching pairs are shown in Fig.~\ref{fig_8} (c).
		
		\item \textbf{The number of cameras} (\# \textit{cam}): 
		The dataset was captured from nine non-overlapping cameras. The layout of the camera network and the example frames from nine cameras are shown in Fig.~\ref{fig_8} (a-b).
		
		\item \textbf{Multiple appearances availability} (\textit{multi-shot}): 
		\texttt{SLP} provides multiple appearances of each person. In this version, we do not provide the original image appearances but provide pre-computed feature descriptors due to legal restrictions. However, we expect that we will provide the original image appearances of people in the future.
		
		\item \textbf{Scene background availability} (\textit{background}): 
		\texttt{SLP} provides the scene background. Although it only contains the background of the camera scene, it contains scene structures such as buildings, ground plane, orthogonal lines etc., which are quite useful for understanding the scene.
		
		\item \textbf{People trajectory availability} (\textit{traj.)}: 
		The ground truth tracking information of every person is available. It includes positions~(x,y locations) and sizes~(height, width) of people in entire image sequences. To annotate and generate people's trajectories for all video sequences, we used an annotation tool `VATIC'~\cite{springerlink:10.1007/s11263-012-0564-1}. The total number of annotated boxes is 1,468,142.
		
		\item \textbf{Camera synchronization} (\textit{sync}):
		Each camera is synchronized with a global timestamp. It provides all timestamps of video frames.
		
		\item \textbf{Camera calibration}: Each camera is calibrated and camera parameters are given.
		
		\item Details of our dataset are summarized in Table~\ref{tab_1}.
		
	\end{itemize}
				
	As mentioned above, our new dataset enables us to test person re-identification methods under practical large-scale surveillance scenarios. Compared to previous datasets, \texttt{SLP} not only provides competitive number of person \textit{IDs} and cameras but also additional information such as full annotations, camera synchronization and camera parameters. 
	In addition, \texttt{SLP} is also applicable to other computer vision tasks such as multi-object tracking, multi-camera activity recognition, and so on.
	It is available on online: \url{https://sites.google.com/view/yjcho/project-pages/re-id_topology}.

	\section{Settings and Methodology}
	\label{sec:eval_meth}
	
	\subsection{Experimental Settings}
	
	Since we mainly focus on person re-identification and camera topology inference problems, we assume that person detection and tracking results are given. 
	Most person re-identification researches follow this assumption and setting.
	In order to evaluate each proposed method, we divided our dataset into two subsets according to time. The first subset contains 1-hour data starting from the global start time~(AM 11:20). It is used in an initialization stage. The latter subset including the remaining data is utilized in an online test stage.   
	
	In this study, every person image is resized to 128$\times$48 pixels.
	We used the LOMO feature~\cite{liao2015person} to describe the appearances of people. 
	It first divides the image into six horizontal stripes. Then it extracts an HSV color histogram and builds descriptors based on Scale Invariant Local Ternary Pattern (SILTP) from each stripe. The final descriptor from the 128$\times$48 image has 26,960 dimensions.
					
	It is worthy to note that the proposed framework does not utilize pre-matched results between cameras.
	In general, many conventional person re-identification methods exploit the pre-matched pairs for learning distance metrics such as LMNN~\cite{weinberger2005distance}, ITML~\cite{davis2007information}, KISSME~\cite{koestinger2012large} and Mahal~\cite{roth2014mahalanobis}.
	The metric learning methods are effective but not appropriate for practical re-identification systems since the matched results are not given at first.
	On the other hand, the proposed framework runs fully automatically, providing the person detection/tracking results of each camera.

	\subsection{Evaluation Methodology}			
	In general, many previous works plot a Cumulative Match Curve~(CMC)~\cite{gray2007evaluating} representing true match being found within the first $n$ ranks for evaluating person re-identification performances. In practice, however, all ranks (2,3,...,$n$) except rank-1 failed to find the correct correspondences. 
	For this reason, rank-1 value is the most important one among all ranks and several studies mainly measure the rank-1 accuracy.
	In this study, we measure the \textit{rank-1 accuracy} by $\frac {TP}{T_{gt}}$, where $TP$ is the number of true matching results and $T_{gt}$ is the total number of ground truth pairs in the camera network. The evaluation metric \textit{rank-1 accuracy} reflects the overall re-identification performance of the camera network.
					
	To evaluate the accuracy of the camera network topology, we adopt two evaluation metrics: \textit{transition time error} and \textit{topology distance}.
	When an inferred transition distribution and a ground truth are given as $(p(\Delta t)$$\sim$$N\left(\mu, \sigma^{2}\right),p_{gt}(\Delta t)$$\sim$$N\left(\mu_{gt}, \sigma_{gt}^{2}\right))$, the \textit{transition time error} is simply measured as $\left|\mu-\mu_{gt}\right|$. 
	We defined the \textit{topology distance} as \textit{Bhattacharyya} the distance that measures the difference between two probability distributions as $d_B(p,p_{gt}){=}-\ln \left(\int \sqrt{p(\Delta t)p_{gt}(\Delta t)}\,\text{d}\Delta t \right)$. If there are multiple links in the camera network, we measure the evaluation metrics for all links and average them to get the final \textit{transition time error} and \textit{topology distance}.    				
				
	\begin{figure}[t]
		\centering
		\subfigure[]{\includegraphics[height=0.27\columnwidth]{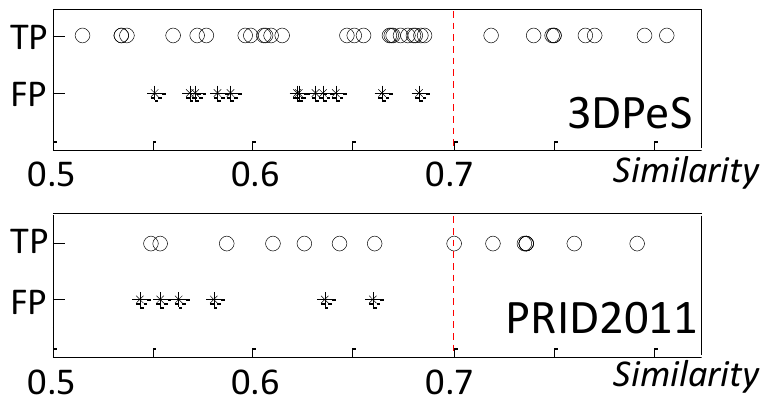}}
		\subfigure[]{\includegraphics[height=0.27\columnwidth]{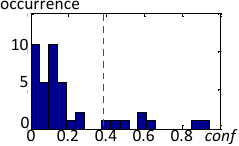}}
		\vspace{-0pt}
		\caption{Observations for selecting reliable parameters}
		\label{fig_scores}
		\vspace{-0pt}
	\end{figure}	
					
	\subsection{Parameters Selection}
	We used fixed parameters for all steps in the Sec.~\ref{sec:proposed} and applied the same parameters to all datasets \texttt{SLP}, \texttt{NLPR MCT}~\cite{Chen2015An}.
	We found that the performance of our framework does not depend on each parameter. The parameters are fixed as follows:
	
	\noindent $\bullet$ $\theta_{sim}$: To determine a proper $\theta_{sim}$, we tested several other datasets (\texttt{3DPeS}~\cite{baltieri2011_308}, \texttt{PRID}~\cite{hirzer11a}) and analyzed the results. Figure~\ref{fig_scores}~(a) shows re-identification similarity scores of each dataset. Marks $\circ$, $\ast$ denote true and false positives.
	To estimate the accurate topology, we need to exclude false positives.
	Based on the test results, we strictly set $\theta_{sim}$ as 0.7 for selecting only true positives.
	
	\noindent $\bullet$ $\theta_{conf}$: Figure~\ref{fig_scores}~(b) shows a histogram of calculated connectivity confidences between nine cameras (36 pairs) in the CAM-to-CAM topology inference step. Weak connections are easily classified, since they have very low confidences ($0-0.25$).
	Based on this, we set $\theta_{conf}$ as 0.4 and fixed the value for topology inference steps and all datasets.
	
	\noindent $\bullet$ Initial $T$: 
	We tested various $T$ such as 300, 450, 600, which are sufficiently wide to find correspondences under unknown camera networks. We noticed that the results of topology inference are not significantly affected by $T$. We set the initial $T$ as 600. 
	
	\noindent * The proposed framework has only one parameter ($T$) to be adjusted depending on the environments. However, $T$ does not critically affect the performance of the proposed system.

	\begin{figure}[t]
		\centering
		\subfigure[{\scriptsize Connectivity confidences}]{\includegraphics[height=0.29\columnwidth]{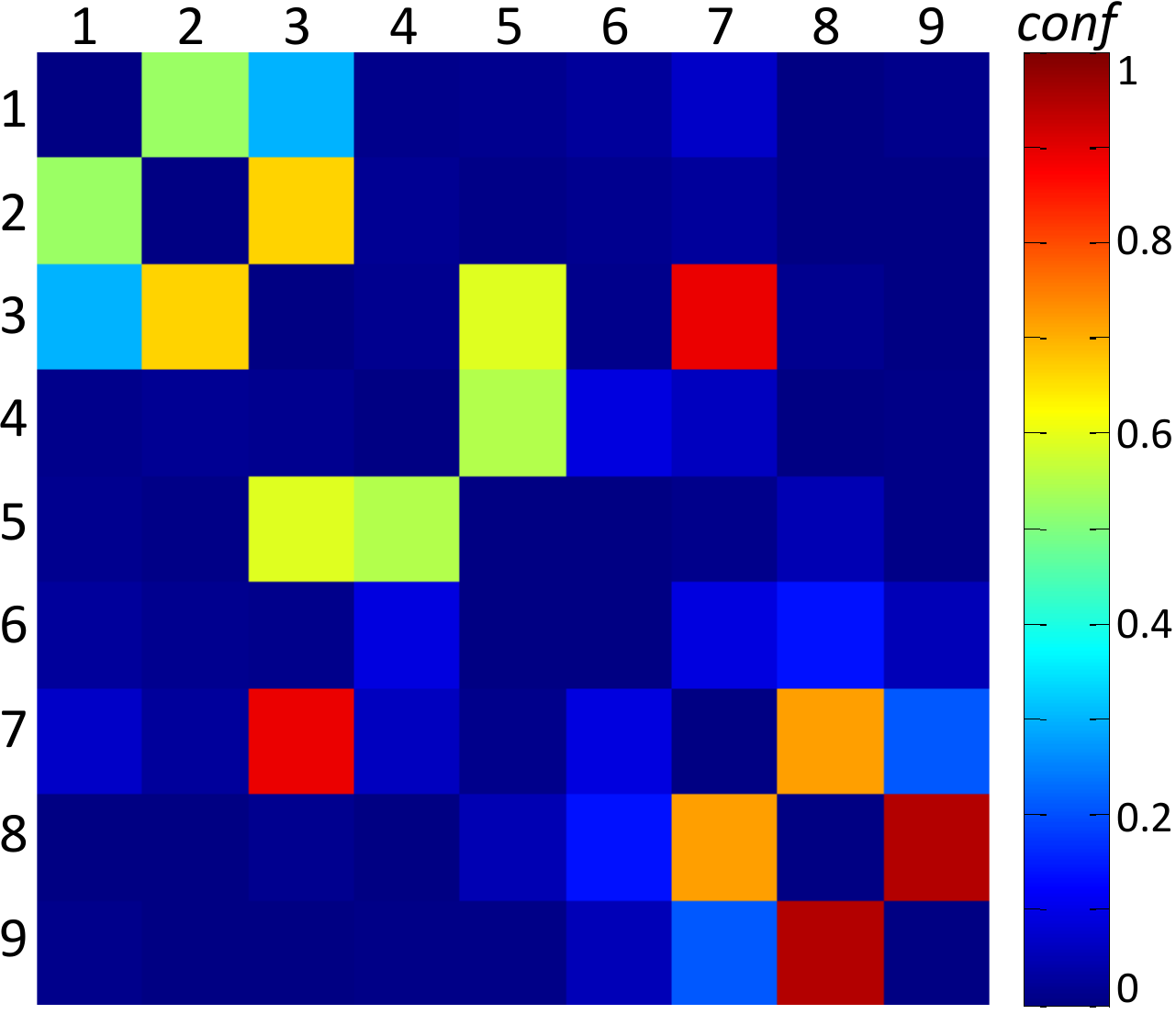}}
		\hspace{5pt}
		\subfigure[Ours]{\includegraphics[height=0.29\columnwidth]{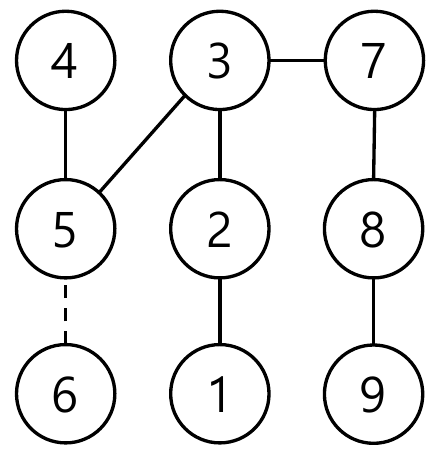}}
		\hspace{5pt}
		\subfigure[DNPR~\cite{martinel2016person}]{\includegraphics[height=0.29\columnwidth]{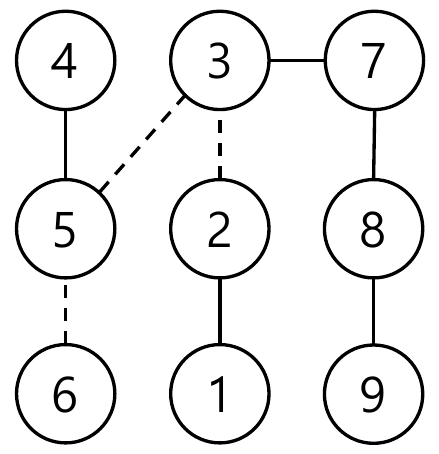}}
		\vspace{-0pt}
		\caption{Results of CAM-to-CAM connectivity check. (a) Connectivity confidence map. (b), (c) Valid camera links of two methods (\textit{solid line}: true link, \textit{dotted line}: missing).}
		\vspace{-0pt}
		\label{fig_10}
	\end{figure}

	\begin{figure}[t]
		\centering
		\subfigure[]{\includegraphics[height=0.38\columnwidth]{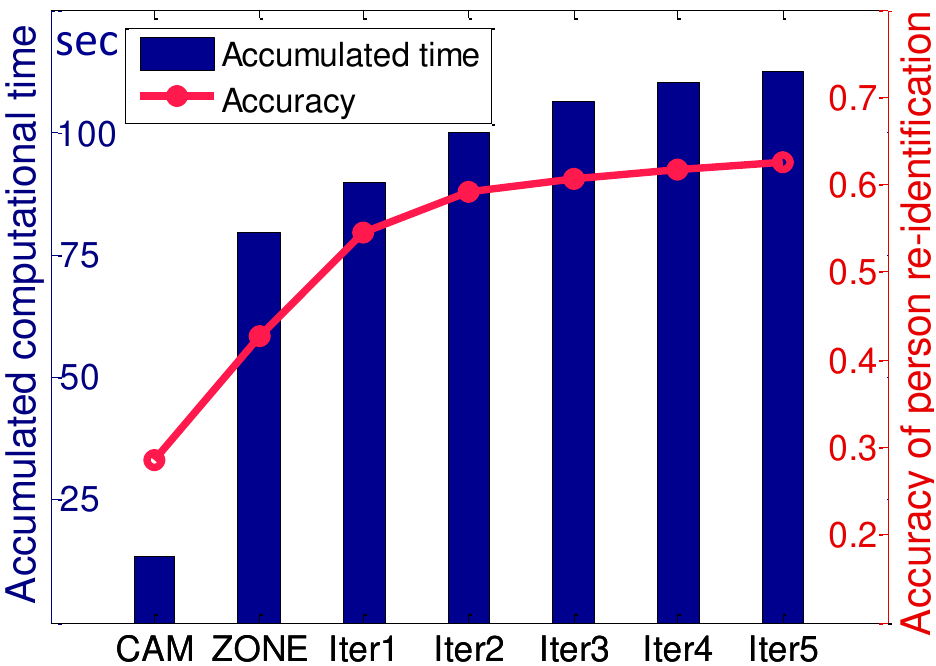}}
		\hspace{10pt}
		\subfigure[]{\includegraphics[height=0.38\columnwidth]{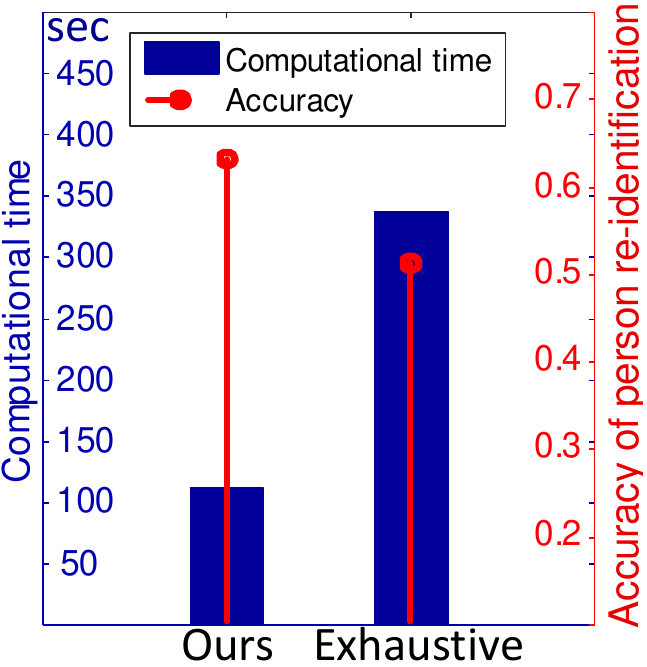}}
		\vspace{-0pt}
		\caption{Results of the person re-identification through the camera topology initialization.}
		\vspace{-0pt}
		\label{fig_11}
	\end{figure}

	\begin{table*}[t]
		\centering
		\caption{Inferred valid Zone-to-Zone links and ground-truths.} \vspace{-0pt}
		\label{tab_4}
		\begin{tabular}{c|c|c|c|c|c||c|c|c|c|c|c}
			\noalign{\hrule height 1pt}	
			Exit & Entry                  & $\mu$  & $\mu_{gt}$ &$\sigma$ & $\sigma_{gt}$ & Exit & Entry &                $\mu$ & $\mu_{gt}$ & $\sigma$ & $\sigma_{gt}$\\ \hline \hline
			\texttt{CAM1-ZONE1}&\texttt{CAM2-ZONE5} & 34.4   &  34.7   & 6.25    & 6.04  & \texttt{CAM2-ZONE5}&\texttt{CAM1-ZONE1} & 40.4  & 40.4   & 7.62  & 5.93 \\ 
			\texttt{CAM2-ZONE2}&\texttt{CAM3-ZONE1} & 36.7   &  36.3   & 8.03    & 5.79  & \texttt{CAM3-ZONE1}&\texttt{CAM2-ZONE2} & 37.6  & 37.0   & 10.3  & 8.90 \\ 
			\texttt{CAM3-ZONE2}&\texttt{CAM5-ZONE6} & -0.42  &  -0.57  & 3.49    & 3.23  & \texttt{CAM5-ZONE6}&\texttt{CAM3-ZONE2} & 0.70  & 1.59   & 3.43  & 2.32 \\  
			\texttt{CAM3-ZONE3}&\texttt{CAM7-ZONE3} & 4.8    &  4.3    & 4.8     & 3.5   & \texttt{CAM7-ZONE3}&\texttt{CAM3-ZONE3} & 3.75  & 4.68   & 2.16  & 3.04 \\ 
			\texttt{CAM4-ZONE4}&\texttt{CAM5-ZONE2} & 30.2   &  30.1   & 13.4    & 12.5  & \texttt{CAM5-ZONE2}&\texttt{CAM4-ZONE4} & 39.5  & 28.6   & 3.82  & 14.8 \\
			\texttt{CAM7-ZONE1}&\texttt{CAM8-ZONE2} & 28.2   &  28.4   & 21.3    & 6.36  & \texttt{CAM8-ZONE2}&\texttt{CAM7-ZONE1} & 31.9  & 30.0   & 2.41  & 4.02 \\ 
			\texttt{CAM8-ZONE1}&\texttt{CAM9-ZONE2} & 11.6   &  11.7   & 4.82    & 4.24  & \texttt{CAM9-ZONE2}&\texttt{CAM8-ZONE1} & 10.5  & 10.5   & 4.03  & 4.08 \\ 
			\noalign{\hrule height 1pt}
		\end{tabular}
		\vspace{-0pt}
	\end{table*}
	
	\begin{figure*}[t]
		\centering
		\subfigure[Initial]{\includegraphics[width=0.6\columnwidth]{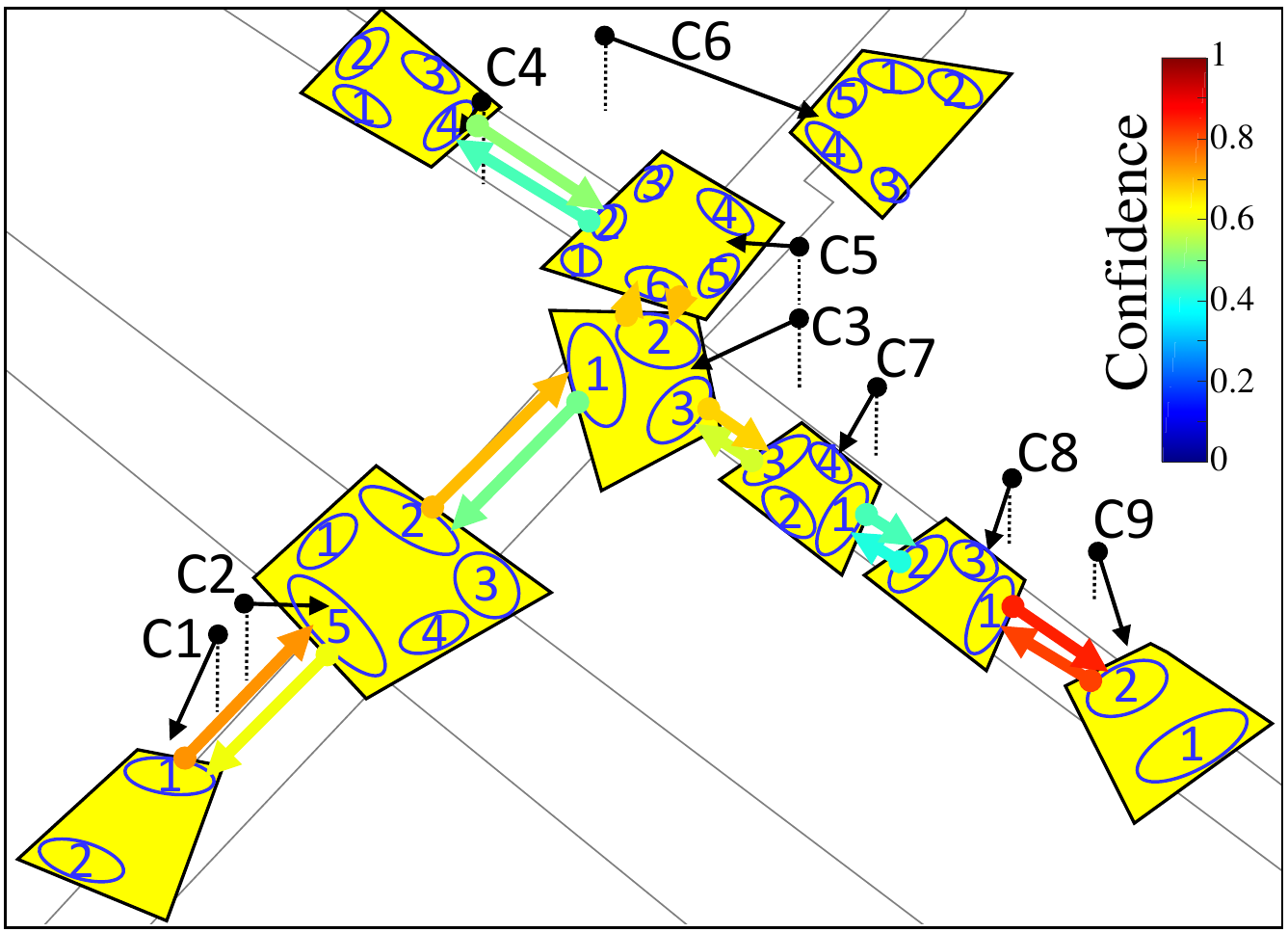}}\hspace{5pt}
		\subfigure[Iteration 1]{\includegraphics[width=0.6\columnwidth]{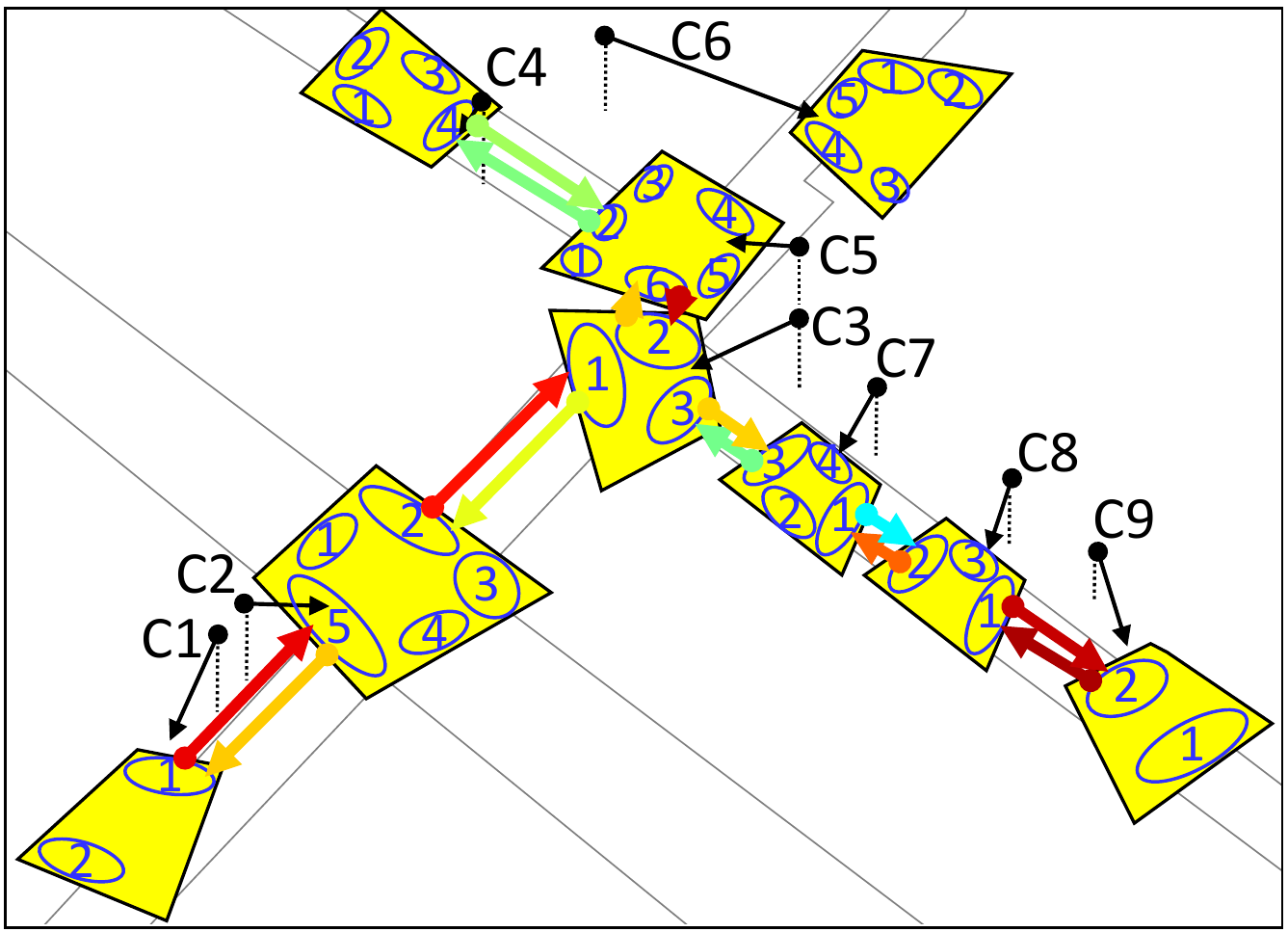}}\hspace{5pt}
		\subfigure[Iteration 5]{\includegraphics[width=0.6\columnwidth]{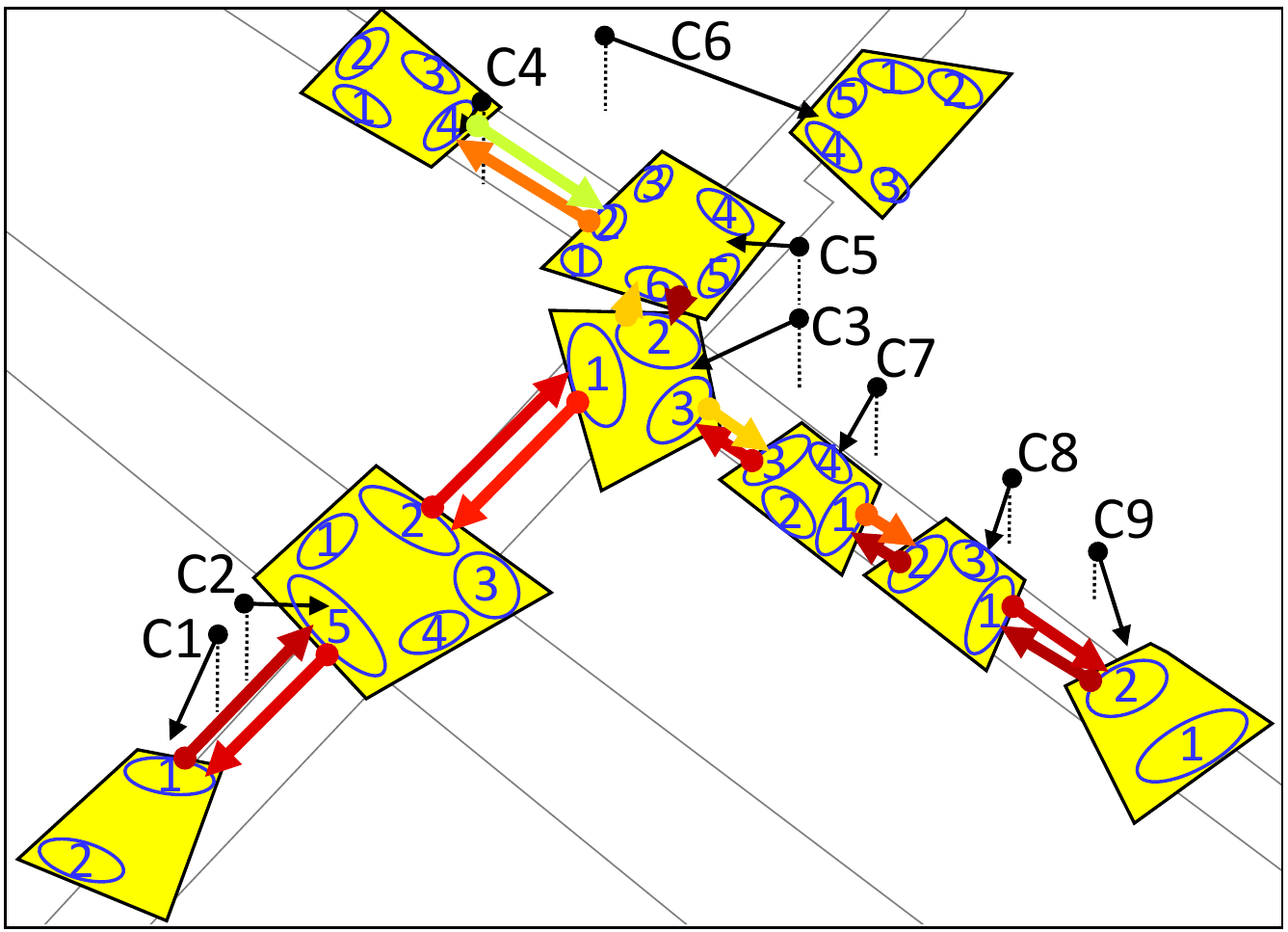}}					
		\caption{The iterative update of connectivity confidences. The connectivity of each link has strengthened during the iterations. (best viewed in color).}
		\label{fig_iteration_con}
	\end{figure*}

	\section{Experimental Results}
	\label{sec:exp}					
	\subsection{Evaluations using Our Dataset: \texttt{SLP}}
	\subsubsection{Camera Network Topology Initialization}
	\label{subsec:exp:cam_to_cam}
	As explained in Sec.~\ref{sec:proposed}, we first inferred the CAM-to-CAM topology between all cameras. We then found the Zone-to-Zone topology and iteratively updated the topology for the topology initialization. During the topology initialization process, we evaluate both tasks: person re-identification and topology inference.

	$\\$		
	\textbf{CAM-to-CAM topology inference result} 			
	
	\begin{table*}[t]
		\centering
		\caption{Performance comparison with other methods in offline initialization stage} \vspace{-0pt}
		\label{tab_3}
		\begin{tabular}{c|c|c|c|c|c|c|c}
			\noalign{\hrule height 1pt}	
			Performances $\backslash$ Methods   & Makris's~\cite{makris2004bridging} & Nui's~\cite{niu2006recovering} & Chen's~\cite{chen2014object} & DNPR~\cite{martinel2016person}   & Cai's~\cite{cai2010recovering}   & LOMO~\cite{liao2015person} -- Exhaustive & Ours   \\ \hline \hline
			Rank-1 accuracy (\%)          & 49.4     & 50.0    & 50.5   & 40.9   & 47.0     & 52.1      & \textbf{62.5} \\ \hline
			Transition time error (sec)  & 28.52    & 23.94   & 16.29  & --     & 4.84    & 11.76     & \textbf{1.21} \\ \noalign{\hrule height 1pt}
		\end{tabular}
		\vspace{-0pt}
	\end{table*}     

	\begin{figure*}[t]
		
		\vspace{-0pt}
		\centering
		\subfigure[Makris~\textit{et al.}~\cite{makris2004bridging}]{\includegraphics[height=0.445\columnwidth]{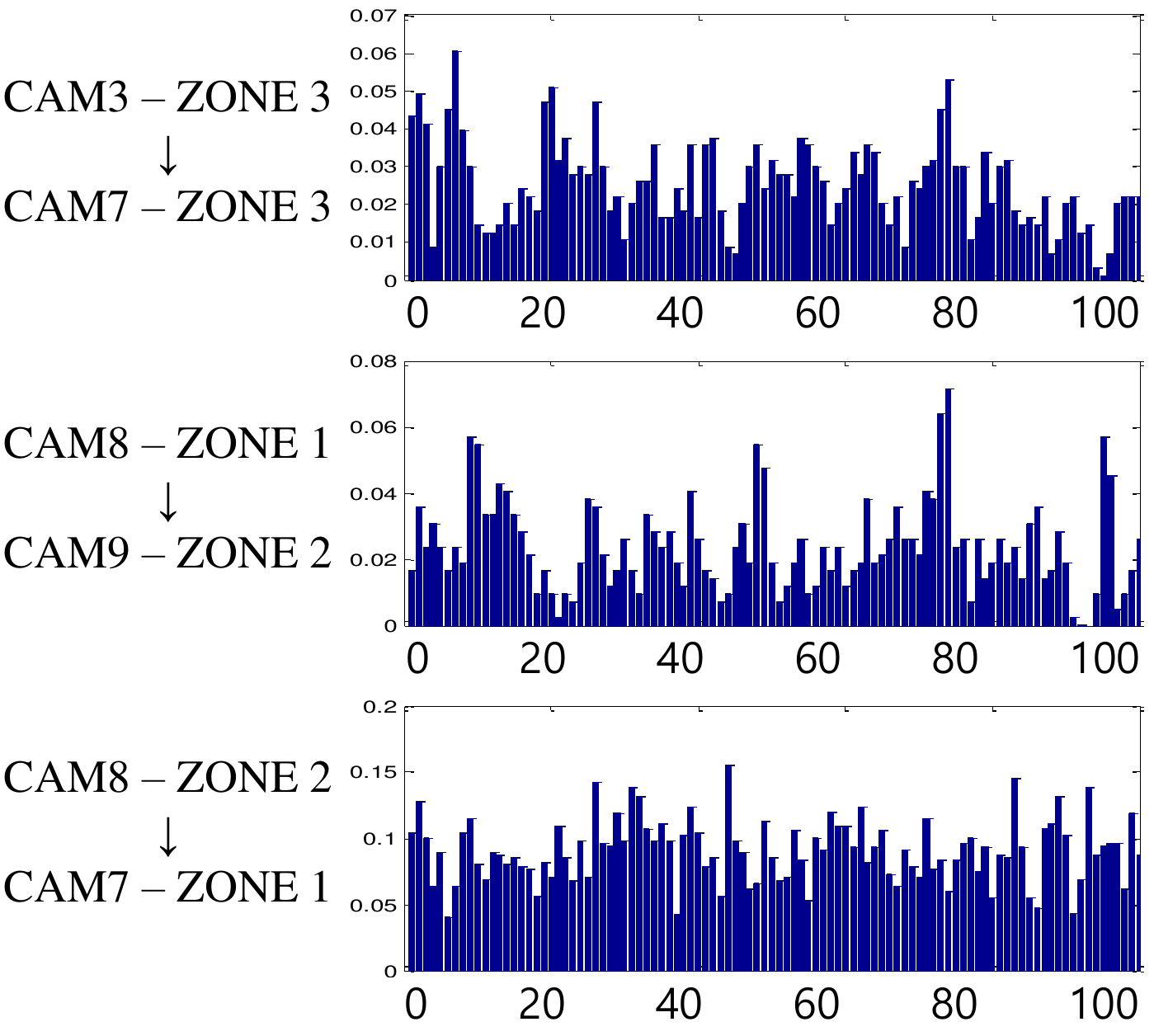}} \hspace{2pt}
		\subfigure[Nui~\textit{et al.}~\cite{niu2006recovering}]    {\includegraphics[height=0.445\columnwidth]{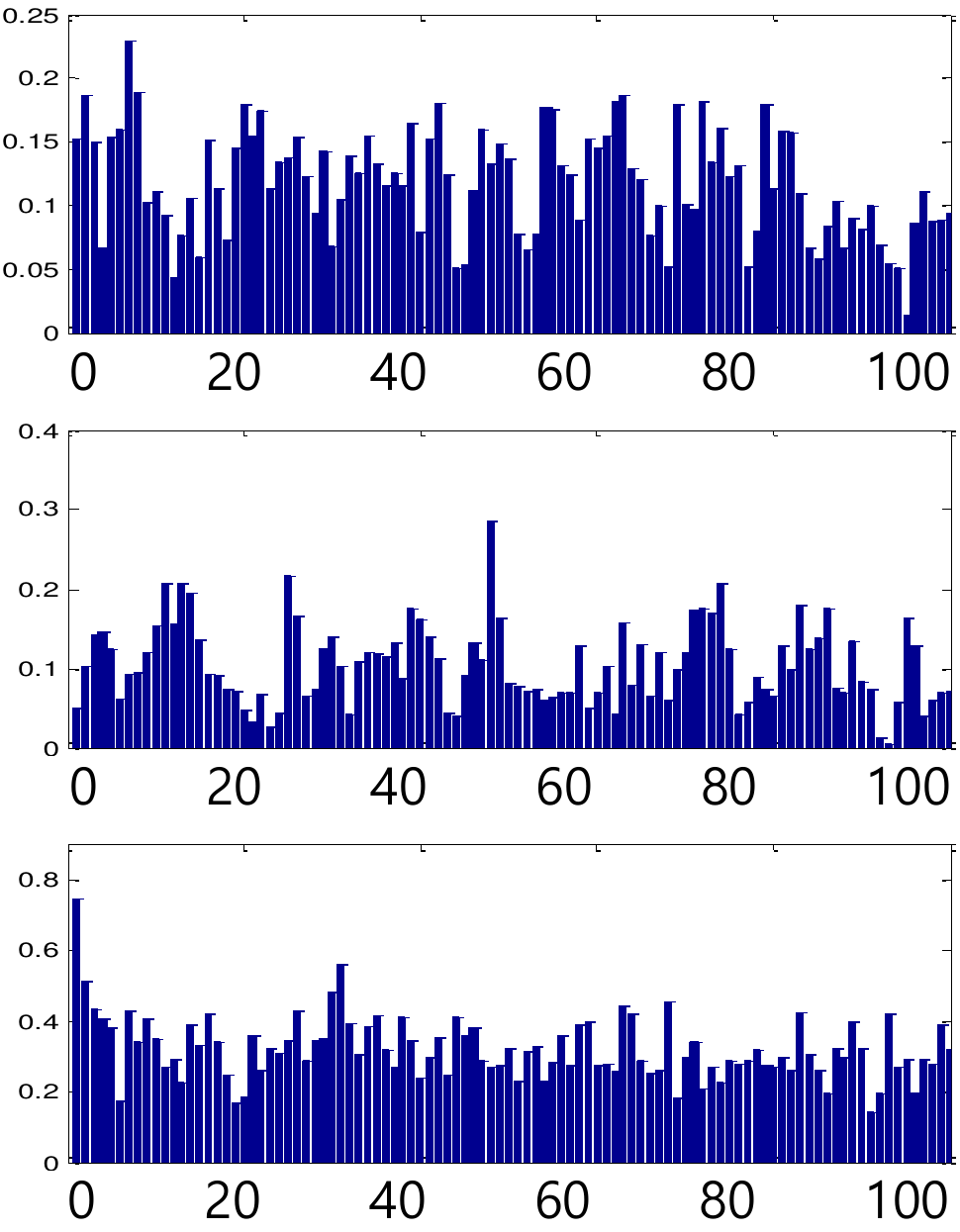}} \hspace{2pt}
		\subfigure[Chen~\textit{et al.}~\cite{chen2014object}]      {\includegraphics[height=0.445\columnwidth]{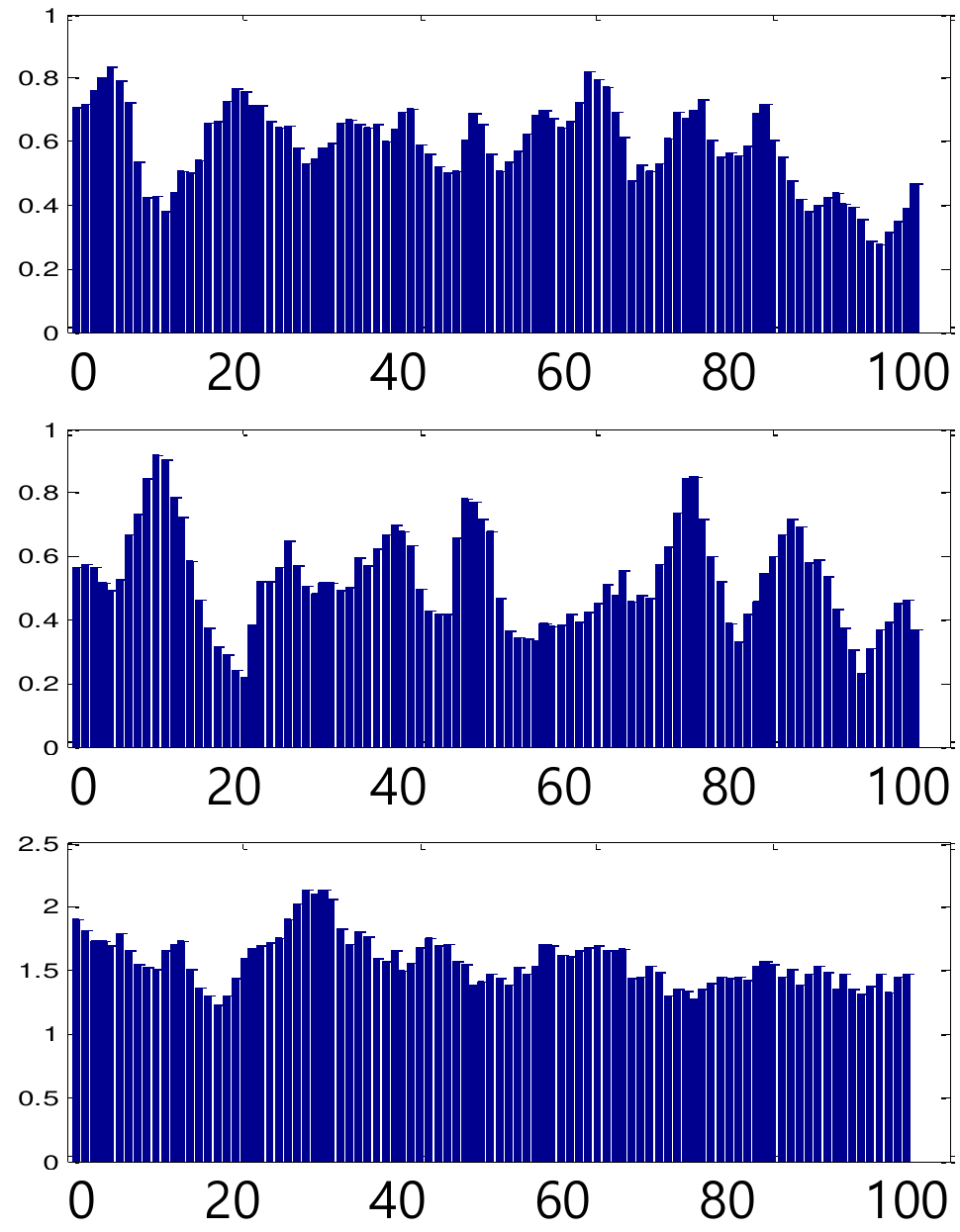}} \hspace{2pt}
		\subfigure[Ours]                                            {\includegraphics[height=0.445\columnwidth]{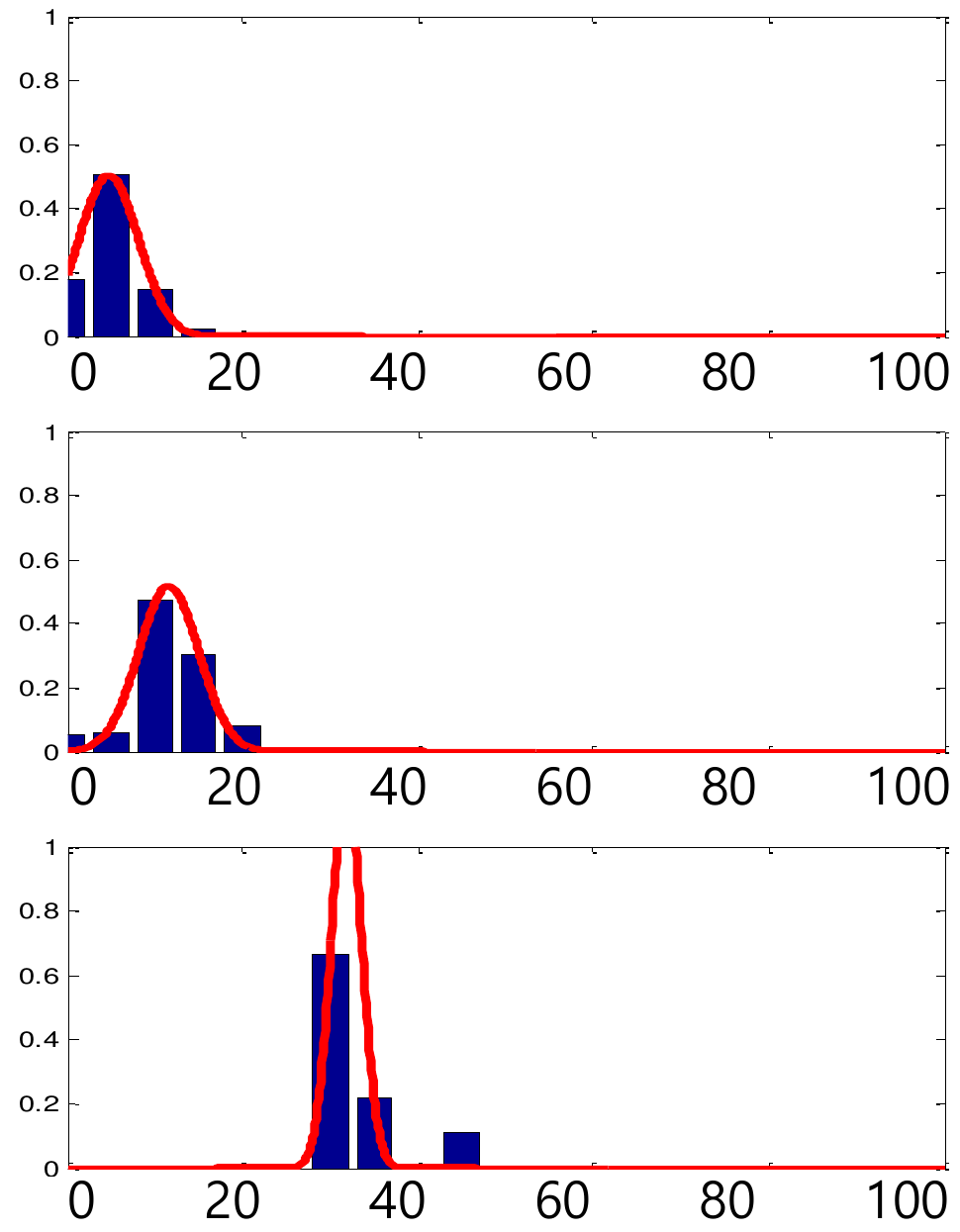}} \hspace{2pt}
		\subfigure[Ground truth]                                    {\includegraphics[height=0.445\columnwidth]{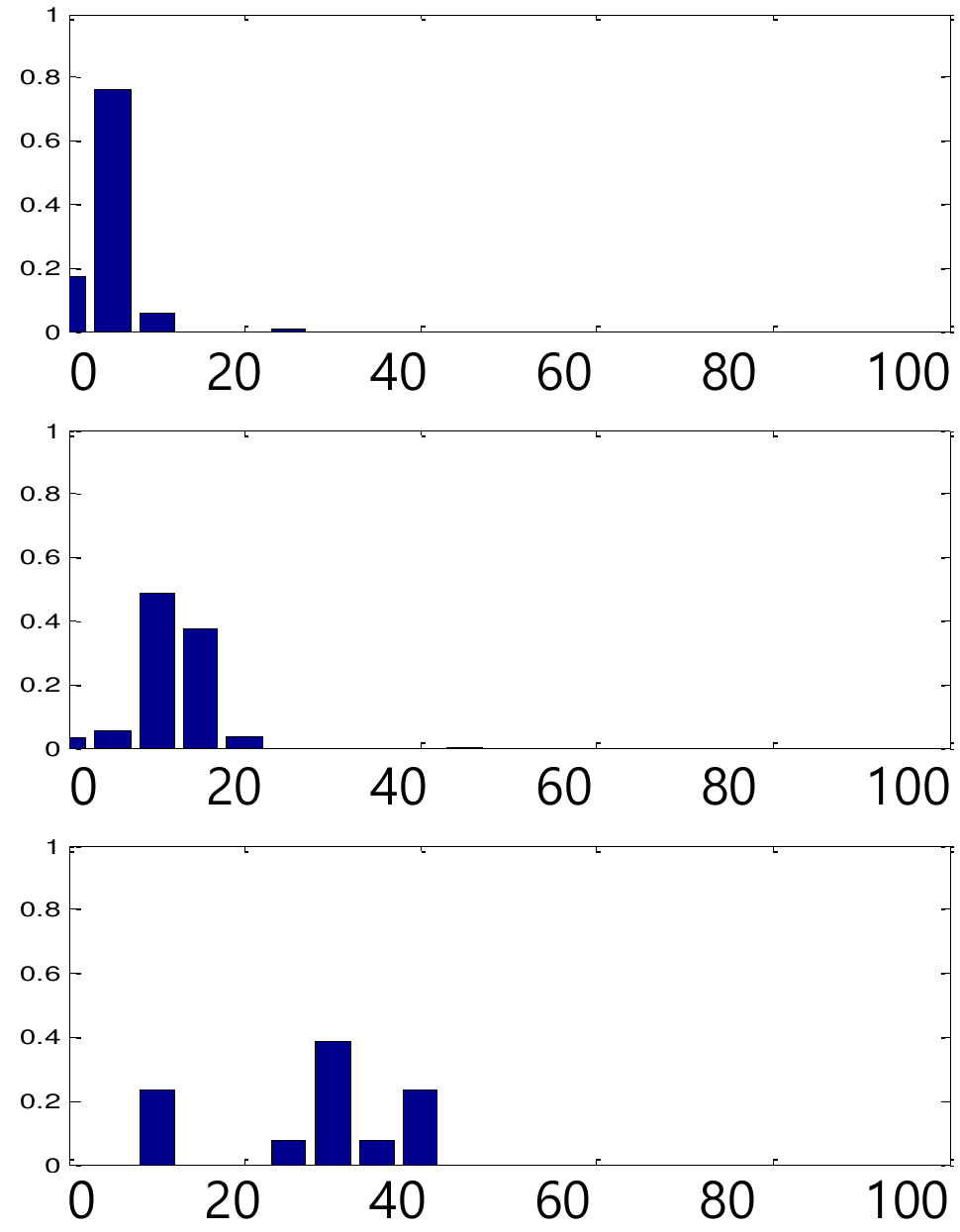}} \\
		\subfigure[Makris~\textit{et al.}~\cite{makris2004bridging}]{\includegraphics[height=0.45\columnwidth]{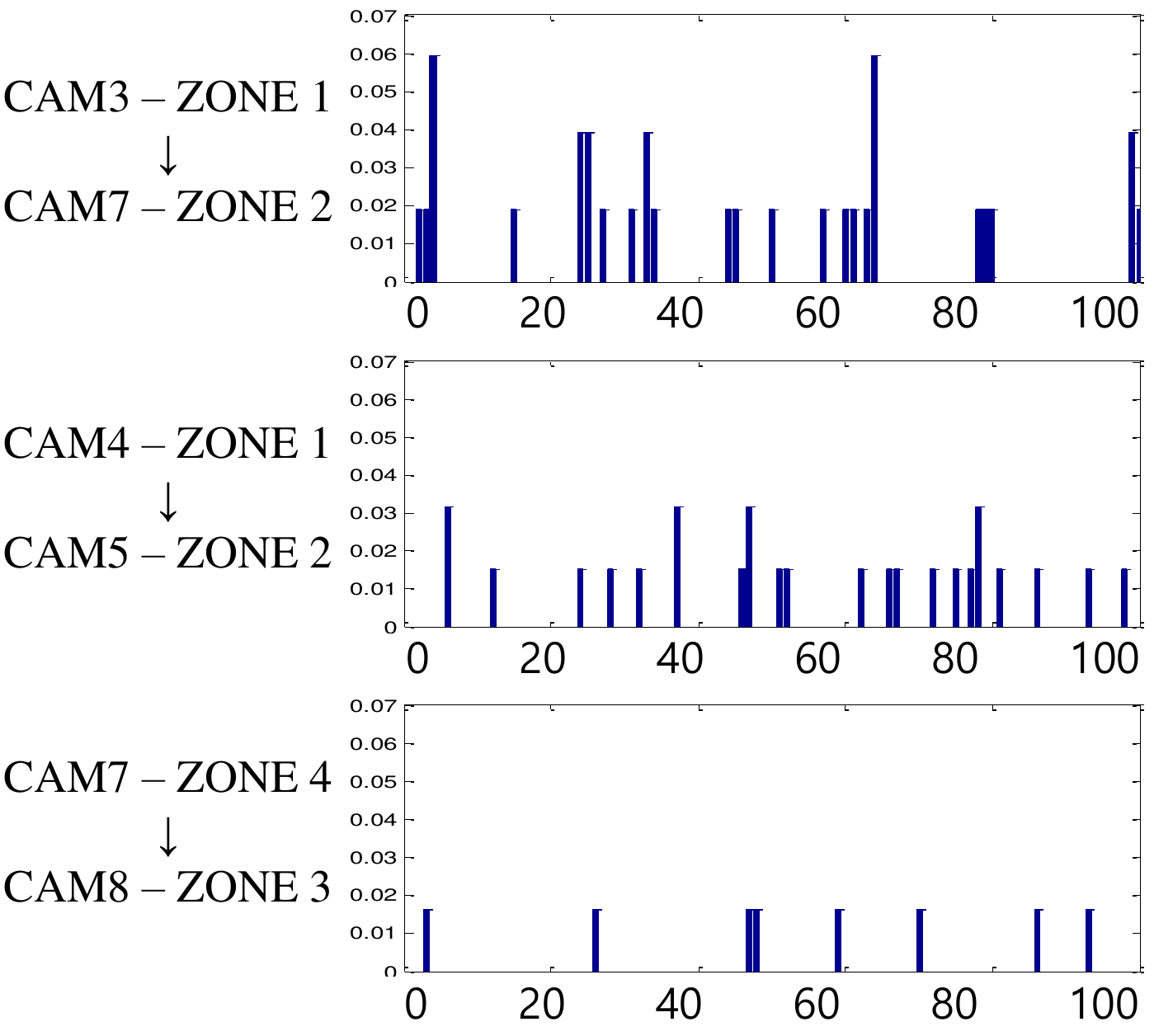}} \hspace{2pt}
		\subfigure[Nui~\textit{et al.}~\cite{niu2006recovering}]    {\includegraphics[height=0.45\columnwidth]{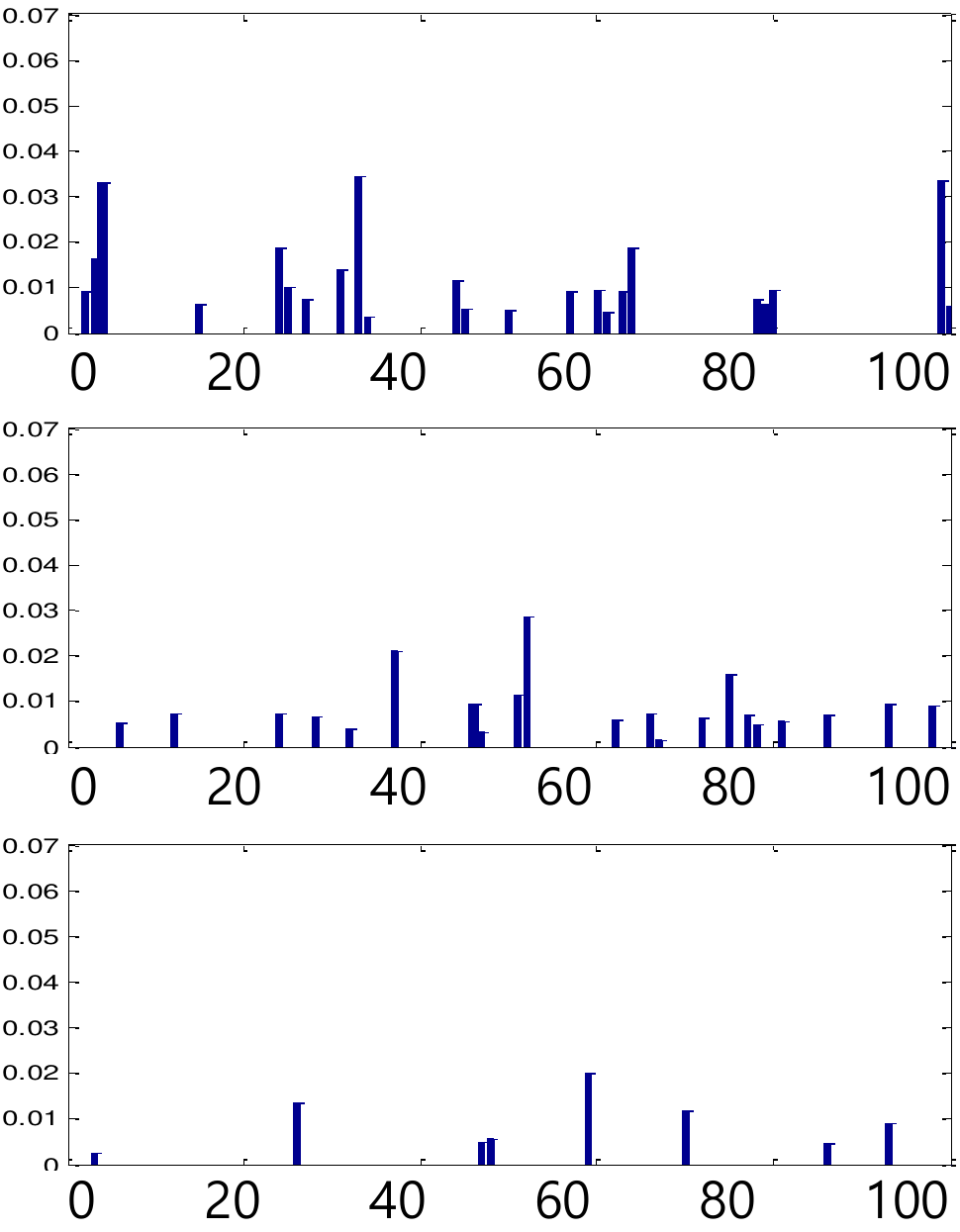}} \hspace{2pt}
		\subfigure[Chen~\textit{et al.}~\cite{chen2014object}]      {\includegraphics[height=0.45\columnwidth]{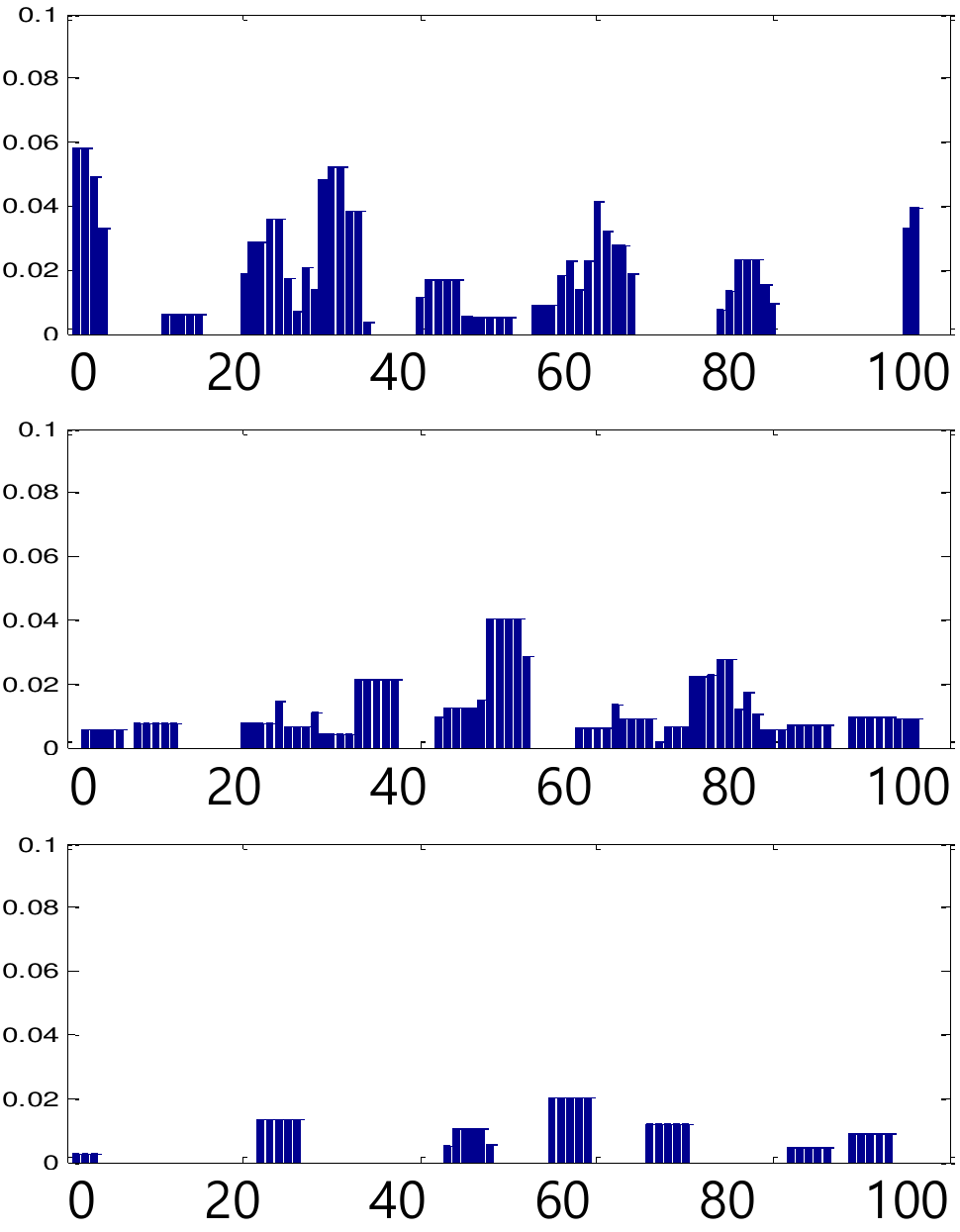}} \hspace{2pt}
		\subfigure[Ours]  											{\includegraphics[height=0.45\columnwidth]{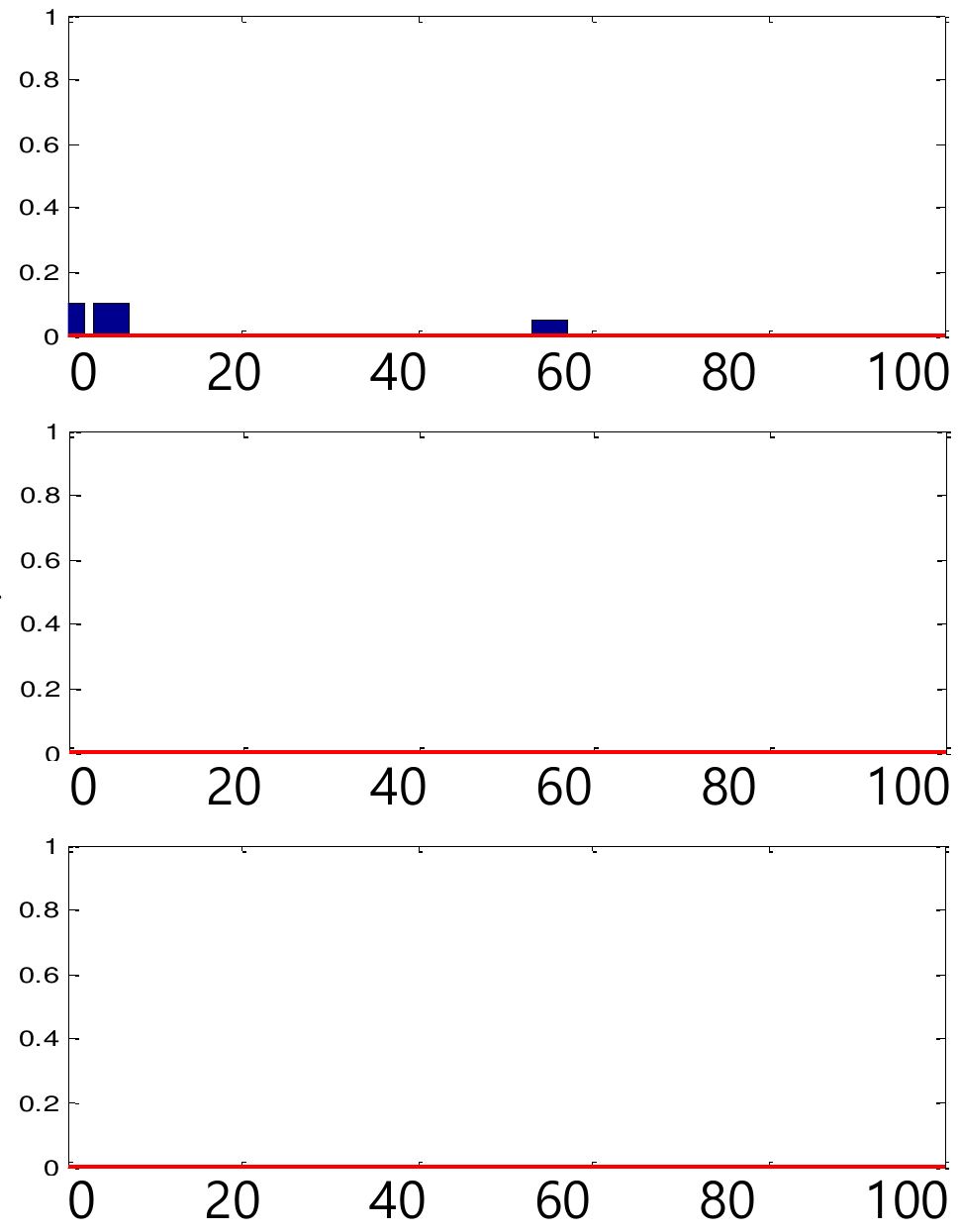}} \hspace{2pt}
		\subfigure[Ground truth]									{\includegraphics[height=0.45\columnwidth]{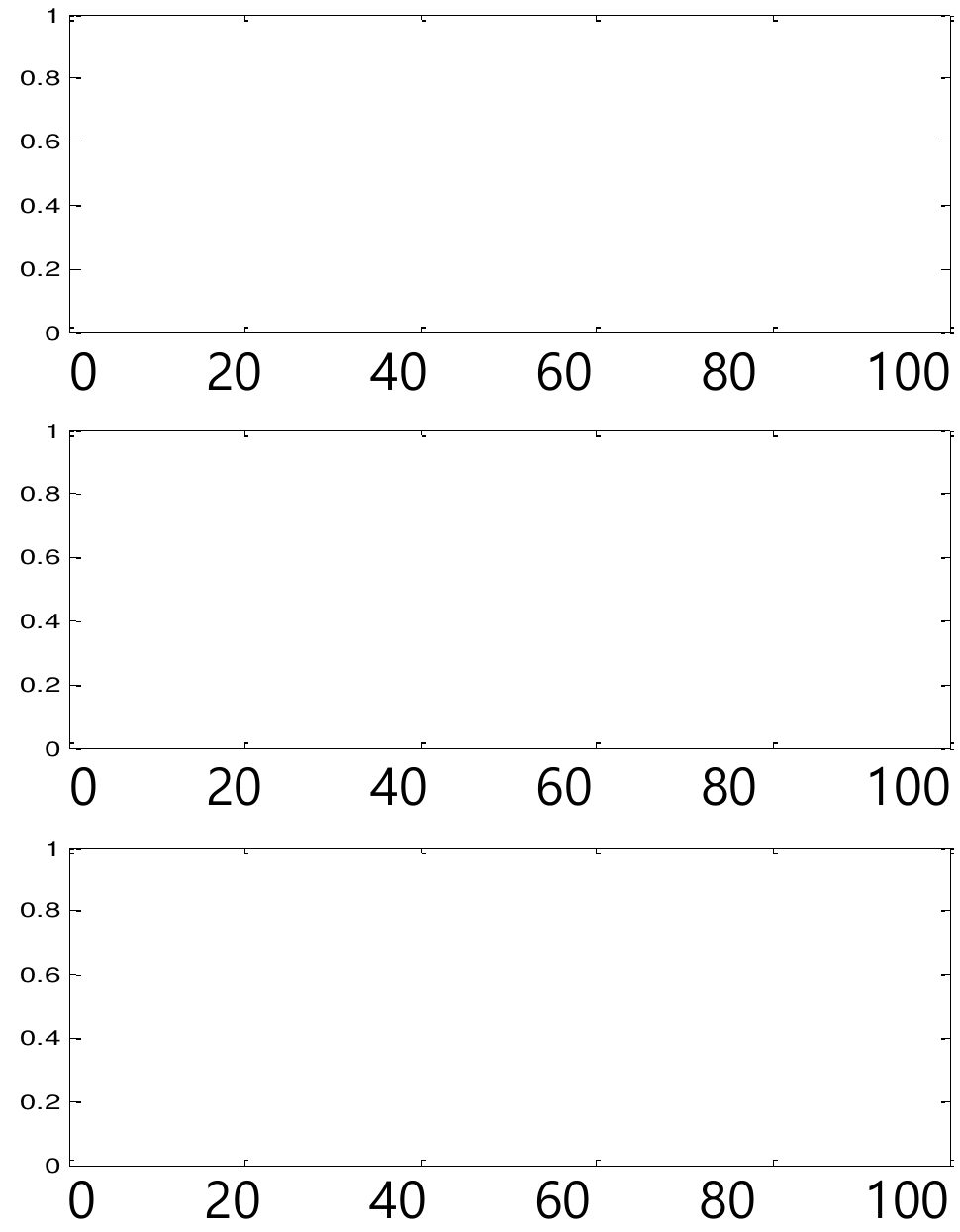}}
		\vspace{-0pt}
		\caption{Comparison of inferred transition distributions and ground truth. First three rows are valid links (a-e) and last three rows are invalid links (f-j).}
		\label{fig_12}
	\end{figure*}

	In order to check the connectivity between two cameras, person re-identification can be carried out bi-directionally.
	However, for the sake of efficiency, we only used uni-directional re-identification results to check the connectivities between the cameras.
	To this end, we assigned a camera containing more people as a training camera (gallery) and a camera containing fewer people as a testing camera (query). That is because the problem of reidentifying a small set of people from a large set is easier than the opposite case. In addition, we can reduce the computational cost for checking the connectivity by half.

	For all the camera pairs, we illustrate a color map of estimated CAM-to-CAM connectivity confidence in Fig.~\ref{fig_10}(a). 
	Each row and column indicates the index of the camera. 
	When the confidence value is greater than $\theta_{conf}$, we regard the corresponding camera pair as a valid link. As a result, the valid camera links are drawn as Fig. \ref{fig_10}(b). Each vertex indicates the index of the camera and valid links are represented by edges. Unfortunately, \texttt{CAM6} failed to be linked to \texttt{CAM5}. That is because the size of the person image patches is very small due to the long distance from the camera. Therefore, it is hard to distinguish the appearances of people. In addition, \texttt{CAM6} is quite isolated from other cameras as we can see from the matching pairs of \texttt{CAM6} in the Fig.~\ref{fig_8}(c). 
	Compared to DNPR~\cite{martinel2016person}, which also infers the CAM-to-CAM topology, our method provides more accurate results. Our proposed method provides an accuracy of 87.5\%, finding 7 links out of 8. The DNPR~\cite{martinel2016person} method provides an accuracy of 62.5\%, finding 5 links out of 8.

	$\\$
	\noindent \textbf{Camera topology initialization results}
	
	Figure~\ref{fig_11} (a) represents the accuracy of person re-identification in each of the proposed initialization steps such as CAM-to-CAM, Zone-to-Zone, and iterative update steps (Sec.~\ref{subsubsec:CAMCAM_con_analy}--\ref{subsubsec:iter_topology_infer}). The accuracy of person re-identification is 28.54\% at the beginning, but it is consistently improved by using inferred and refined camera topology information.
	As a result, our method reaches 62.55\% accuracy at the last step of the topology initialization. 
	In addition, it took only 112.46 seconds to conduct both person re-identification and camera topology inference tasks with a large number of people in the nine cameras (using Intel i7 CPU in MATLAB).
	Figure~\ref{fig_iteration_con} shows the connectivity confidences between the linked camera pairs. The connections become stronger during the iterations. In the final iteration, most connections demonstrate a high connectivity confidence ($conf\left( p\left(\Delta t\right) \right) > 0.7$). 
	On the other hand, a connection from Exit (\texttt{CAM4-Zone4}) to Entry (\texttt{CAM5-Zone2}) is not a strong connection, since a small number of people moved from \texttt{CAM4} to \texttt{CAM5} during the topology initialization period (AM 11:20 -- PM 12:20).
	
	Figure~\ref{fig_11} (b) shows the comparison of our proposed approach with a conventional approach called an ”Exhaustive” approach, which fully compares the multiple appearances of people and exhaustively searches the correspondences of people between the entry/exit zones without using camera topology information.
	For the baseline of the exhaustive method, we used a LOMO feature~\cite{liao2015person} descriptor.
	It shows a low performance (52.06\% person re-id accuracy) compared to the proposed method. Moreover, it takes a much longer time (337.27 seconds) than our proposed method.	
								
	A list of valid Zone-to-Zone links inferred by the proposed methods is summarized in Table.~\ref{tab_4}.
	Note that the pair of Exit (\texttt{CAM3-ZONE2}) -- Entry (\texttt{CAM5-ZONE6}) has the negative value of the transition time because these zones are overlapped. The overall results are close to the ground-truth $N\left( \mu_{gt},\sigma_{gt}^{2} \right)$.
	We expect that the inferred transition distributions of valid links show Gaussian-like distributions and those of invalid links show uniform or arbitrary distributions.
	However, the previous methods~\cite{makris2004bridging, niu2006recovering, chen2014object} showed unclear and noisy distributions for both valid and invalid links as shown in Fig.~\ref{fig_12} (a-c,f-h).
	On the other hand, our results are very similar to the ground truth (Fig.~\ref{fig_12} (d,e,i,j)).
	
	In Table.~\ref{tab_3}, we compared several methods in terms of re-identification and topology accuracies. 
	To ensure a fair comparison of the previous methods with our method, we applied the same baseline (e.g. feature descriptor, parameters, experimental settings) to the previous methods except topology inference methods.
	The inferred topologies of previous methods~\cite{makris2004bridging, niu2006recovering, chen2014object} are used for person re-identification tasks by following the method in \cite{loy2010time}. 
	Note that we cannot estimate the \textit{transition time error} of DNPR~\cite{martinel2016person} since the method only infers CAM-to-CAM connections.
	As we can see, our method shows superior performances for both re-identification and topology inference than other methods.

	\begin{table*}[t]
		\centering
		\caption{Performance comparison in online test stage.} \vspace{-0pt}
		\label{tab_5}
		\begin{tabular}{l|c|c|c|c}
			\noalign{\hrule height 1pt}
			\multirow{2}{*}{}       & \multicolumn{2}{c|}{\footnotesize Offline initialization stage} & \multicolumn{2}{c}{ Online test stage}  \\ \cline{2-5} 
			& {\footnotesize Rank-1 accuracy}  & {\footnotesize Topology dist} & {\footnotesize Rank-1 accuracy} & {\footnotesize Topology dist}   \\ \hline\hline
			LOMO\cite{liao2015person} -- Exhaustive    & 52.1\%                    & 5.620                  & 65.6\%              & 0.046                  \\ \hline
			Ours -- no update      & \multirow{2}{*}{62.5\%}   & \multirow{2}{*}{0.076} & 72.3\%              & 0.093                  \\ \cline{1-1} \cline{4-5} 
			Ours                   &                           &                        & \textbf{74.8\%}     & \textbf{0.023}         \\ \hline \hline
			True matching & 100\%                     & 0                      & 75.6\%              & 0.011                  \\ \noalign{\hrule height 1pt}	
		\end{tabular}
		\vspace{-0pt}
	\end{table*}				
				
	\subsubsection{Online Person Re-identification Results}
	\label{subsec:exp:online}
	
	Based on the inferred camera topology in the initialization stage, we conducted online person re-identification and compared it with two different approaches.
	The first approach estimates the camera topology based on \textit{Exhaustive search} in the offline initialization stage and uses the inferred topology for the online re-identification test. Note that this approach exploits the inferred topology but still fully compares multiple appearances of people to find correspondences in the online test stage.
	The second approach based on \textit{True matching} estimates the camera topology using ground truth re-identification pairs in the offline initialization stage and performs the online re-identification test in the same way as our method. 
	
	As shown in Table~\ref{tab_5}, the performance of our online re-identification test is comparable to that of \textit{True matching}. 
	Our method also outperforms \textit{Exhaustive search} in terms of both re-identification accuracy and topology distance. 
	Thanks to the online topology update in the online re-identification test stage, the topology distance of our method decreased during the online test stage { (0.076 $\rightarrow$ 0.023)}.
	As we can see in the table, our method with the online topology update (\textit{i.e., ours}) performs better than our method without online topology update (\textit{i.e., Ours-- no update}) in terms of both re-identification accuracy and topology distance in the online test stage. 
	It demonstrates that our online topology update and re-identification methods are effective and complement each other.

	\subsection{Evaluations using Public Dataset: \texttt{NLPR MCT}}
	\label{subsec:exp:MCTdata}

\begin{figure}[t]
		\centering
		\subfigure[A layout of the dataset]{\includegraphics[width=0.8\columnwidth]{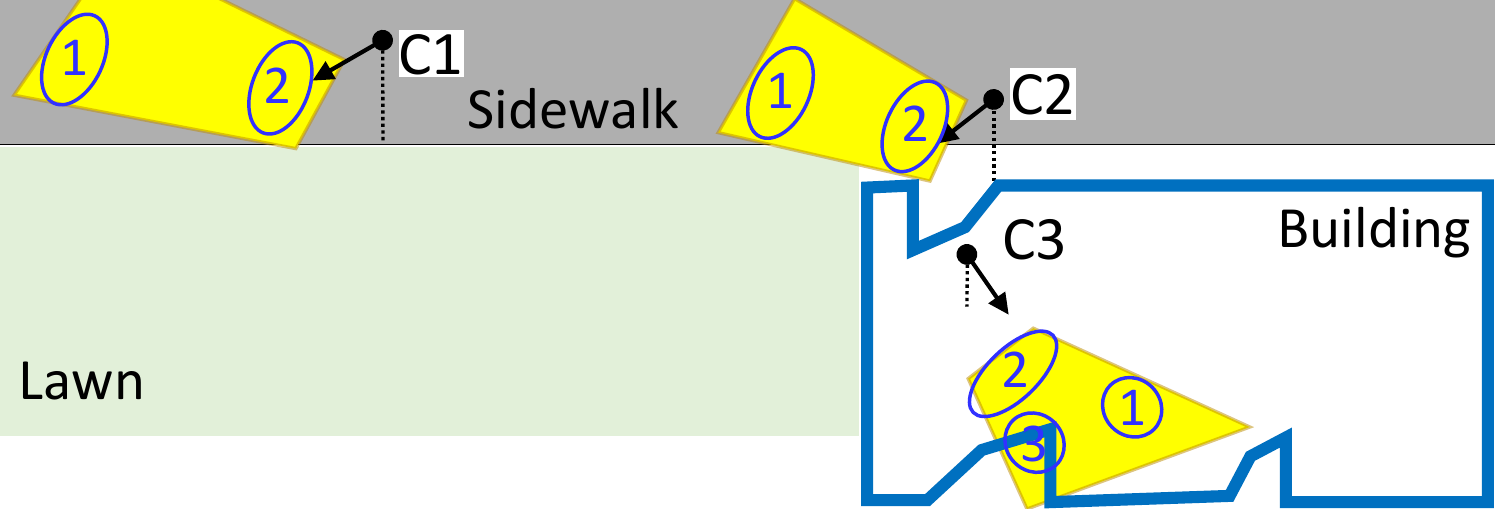}} \hspace{5pt}
		\subfigure[CAM1]{\includegraphics[width=0.3\columnwidth]{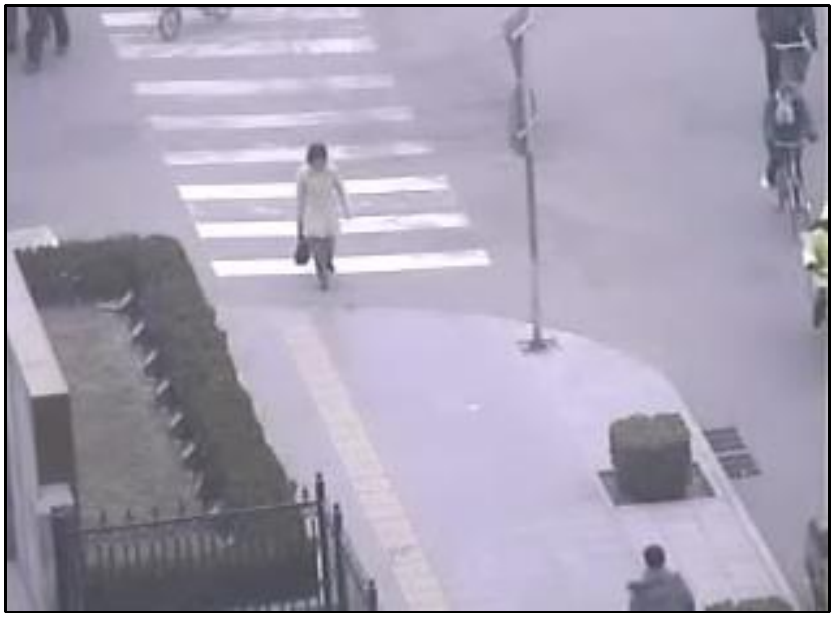}}
		\subfigure[CAM2]{\includegraphics[width=0.3\columnwidth]{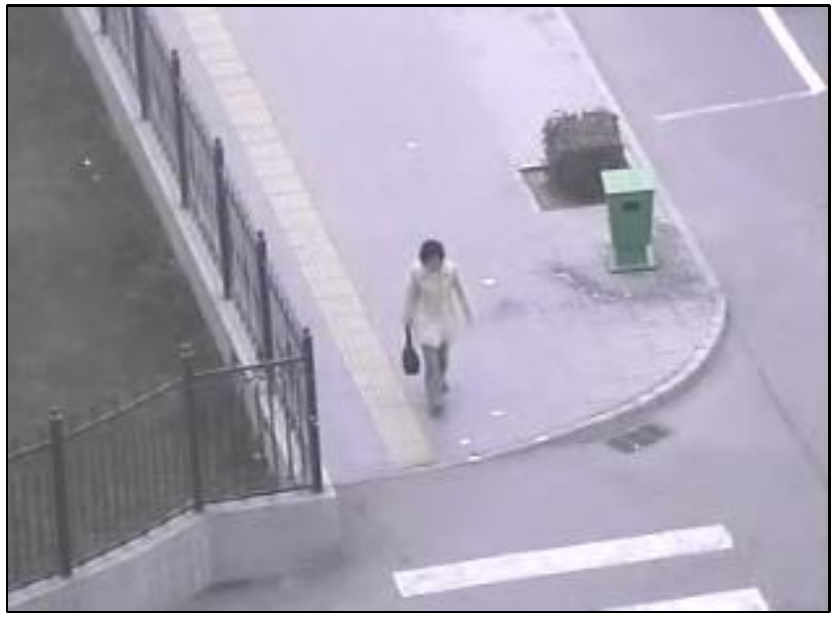}}
		\subfigure[CAM3]{\includegraphics[width=0.3\columnwidth]{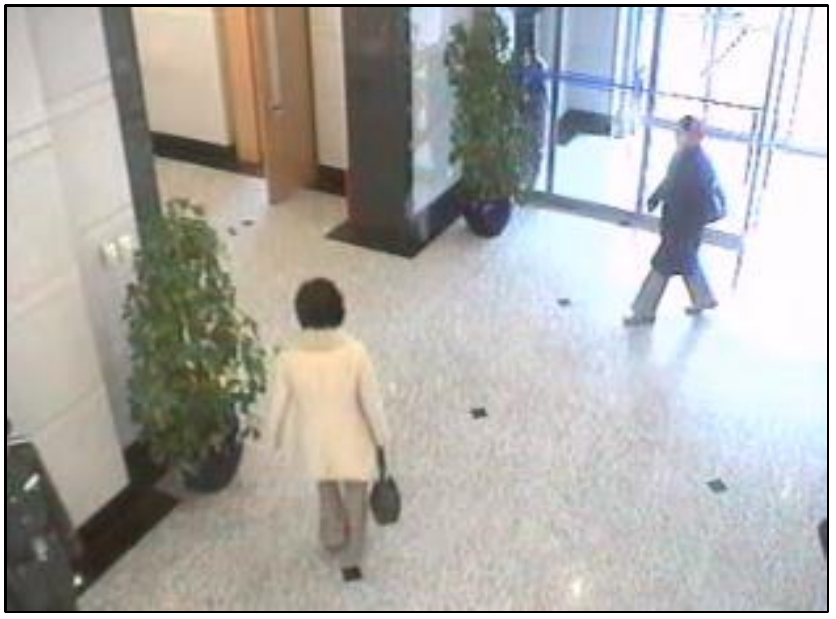}}
		\caption{A public multi-camera person re-identification dataset: \texttt{NLPR MCT}~\cite{Chen2015An}}
		\vspace{-4pt}
		\label{fig_public_data}
	\end{figure}

	In order to validate the proposed method using public datasets, we tested the proposed method using the \texttt{NLPR MCT}~\cite{Chen2015An} dataset.
	The dataset contains four subsets with different non-overlapping multi-camera networks and different numbers of people. Among them, we used two datasets (i.e., DATA1, DATA2) containing larger numbers of people than others (the number of people -- DATA1: 235, DATA2: 255, DATA3: 14, DATA4: 49).
	DATA1 and DATA2 have the same layout and camera field of views as shown in Fig.~\ref{fig_public_data} (a, b-d).
	Duration of each dataset is 20 minutes. 
	Since the datasets are not large-scale datasets (the number of people is small), we used whole videos (20min) for our proposed camera network topology initialization method and did not perform the online person re-identification.

	$\\$
	\noindent \textbf{CAM-to-CAM topology inference}
	
	Based on our proposed method, we inferred the CAM-to-CAM topologies of both datasets (DATA1, 2).
	We found strong connections between \texttt{CAM1} and \texttt{CAM2} in both datasets. Unfortunately, we could not find the connections from \texttt{CAM2} to \texttt{CAM3} in both datasets, although there is a uni-directional link from Exit (\texttt{CAM2-Zone2}) to Entry (\texttt{CAM3-Zone2}) because of the severe illumination difference (indoor and outdoor) and different radiometric parameters between the two cameras. 
	
	In this case, person re-identification methods based on matching should handle these photometric issues, which is beyond the scope of this study. In this case, even the exhaustive search approaches failed to find the connection between \texttt{CAM2} and \texttt{CAM3} because of the same reasons.
	We expect that by adopting some color correction methods such as \cite{chen2014object} before re-identification or exploiting illumination invariant feature descriptors for re-identification, we can find the connection between \texttt{CAM2} and \texttt{CAM3}. We will try this in future studies.

	\begin{figure*}[t]
		\vspace{-5pt}
		\centering
		\subfigure[Makris~\textit{et al.}~\cite{makris2004bridging}]{\includegraphics[height=0.21\columnwidth]{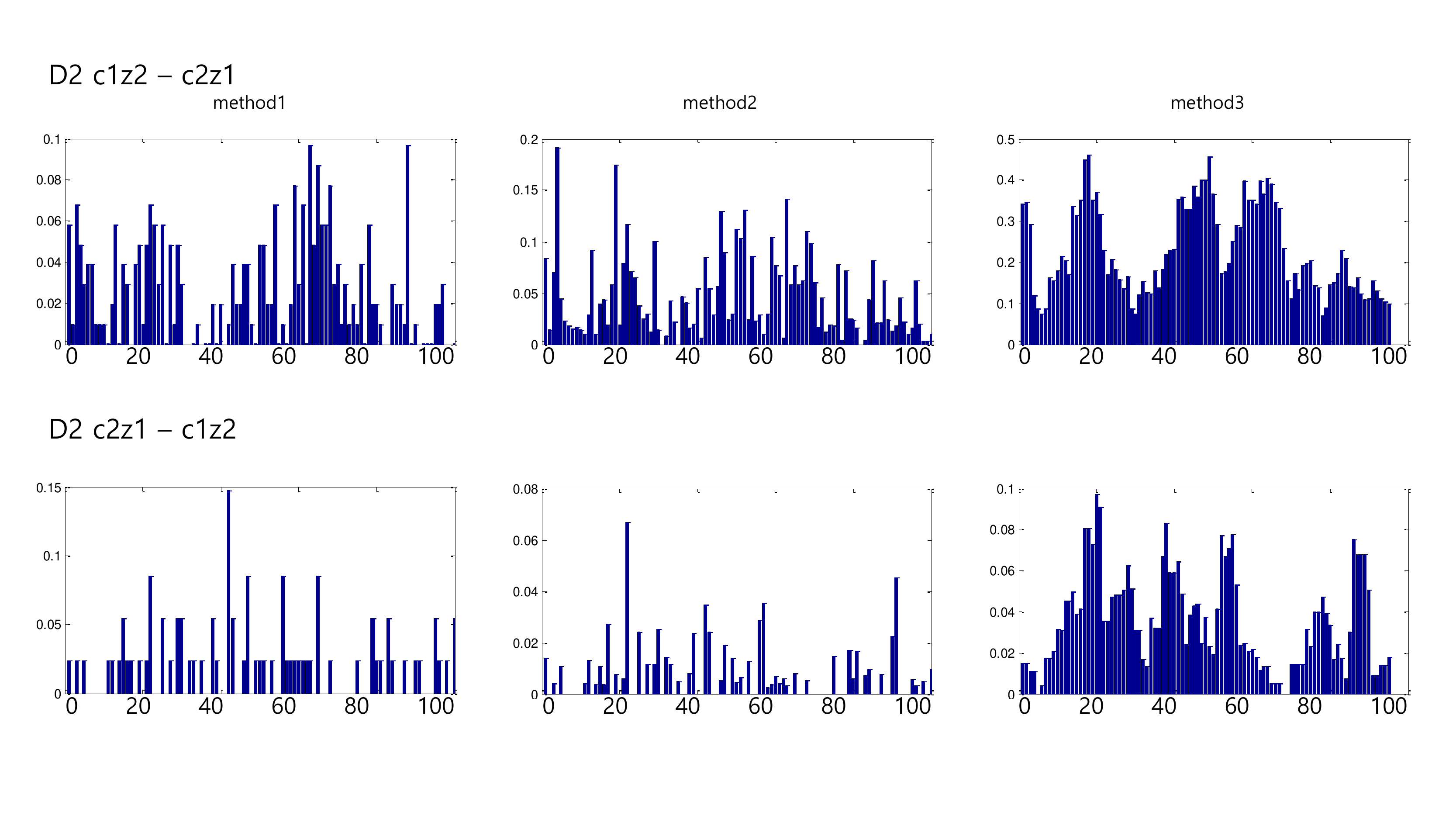}} \hspace{1pt}
		\subfigure[Nui~\textit{et al.}~\cite{niu2006recovering}]{\includegraphics[height=0.21\columnwidth]{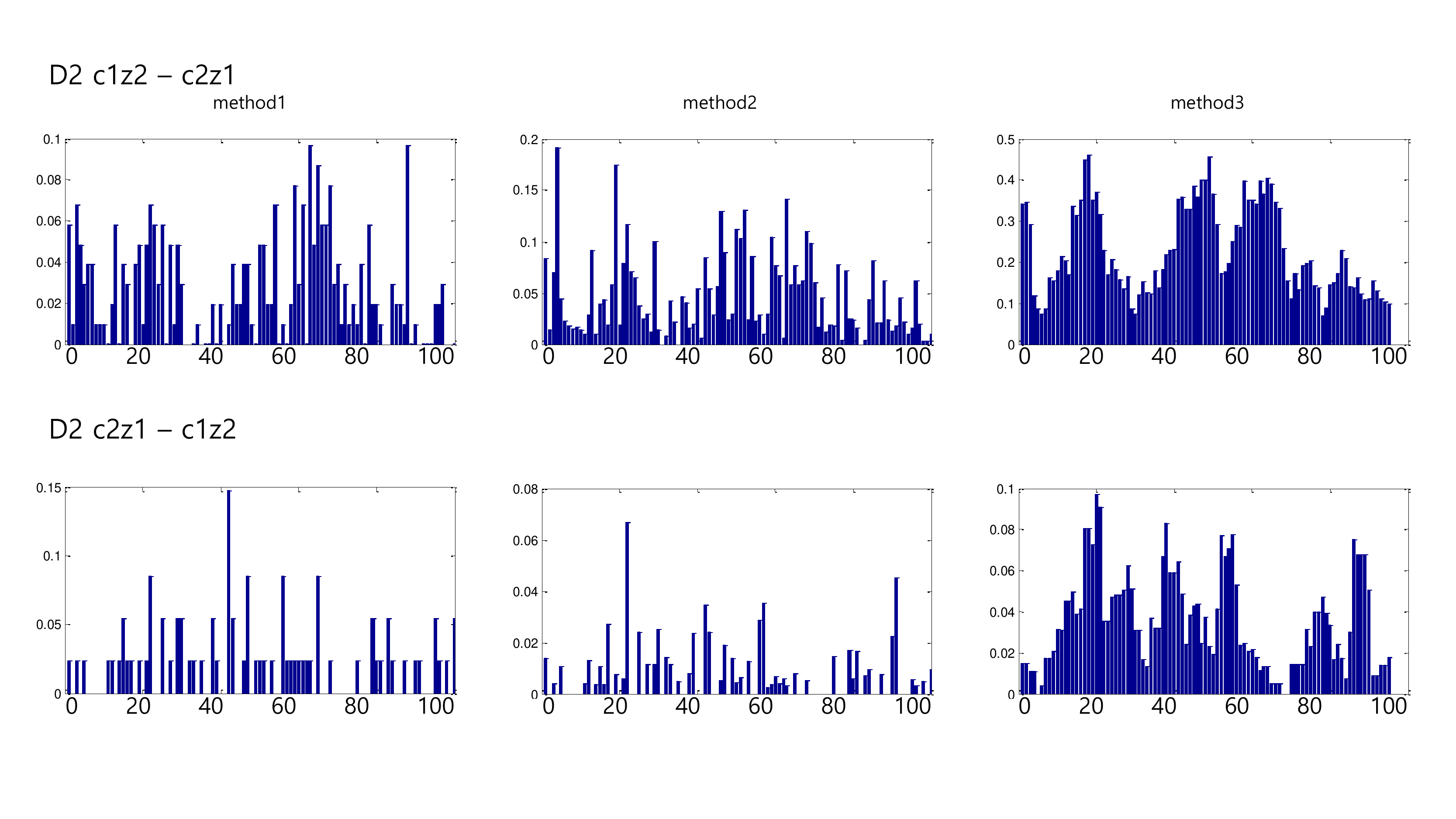}} \hspace{1pt}
		\subfigure[Chen~\textit{et al.}~\cite{chen2014object}]{\includegraphics[height=0.21\columnwidth]{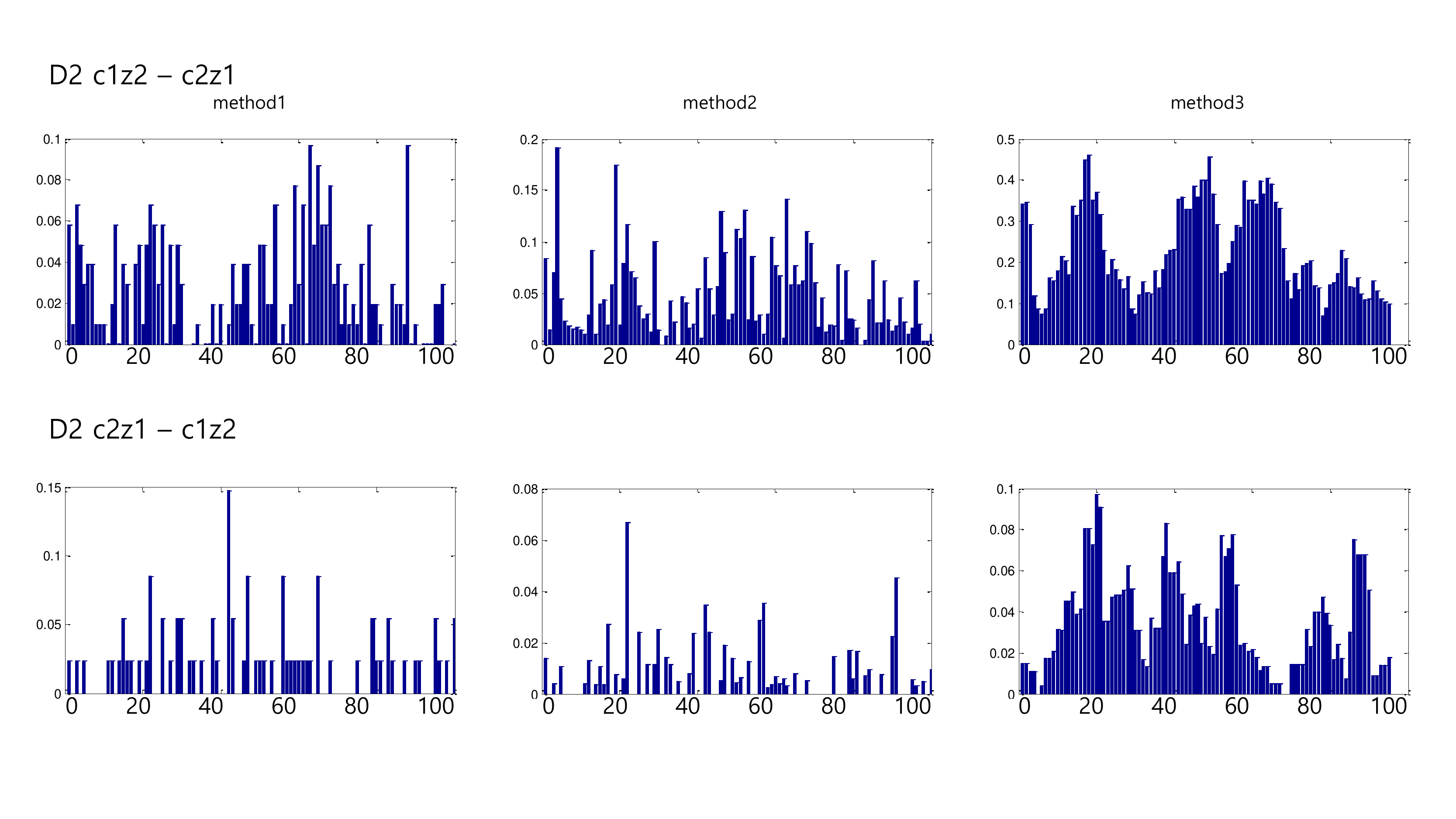}} \hspace{1pt}
		\subfigure[Ours]  {\includegraphics[height=0.21\columnwidth]{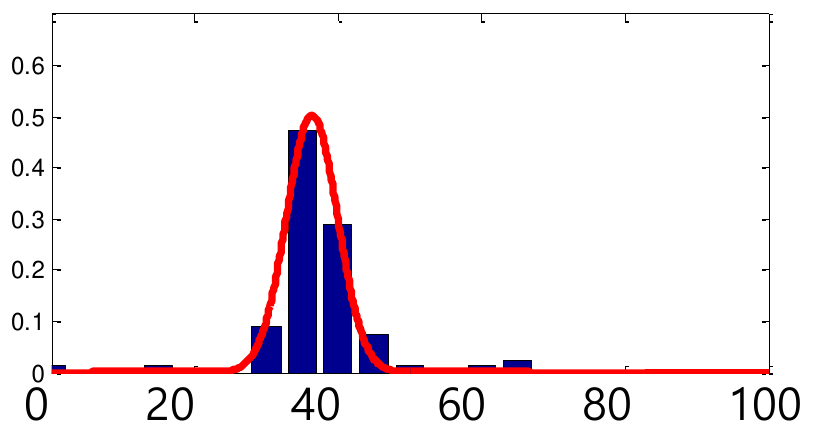}} \hspace{1pt}
		\subfigure[Ground truth]{\includegraphics[height=0.21\columnwidth]{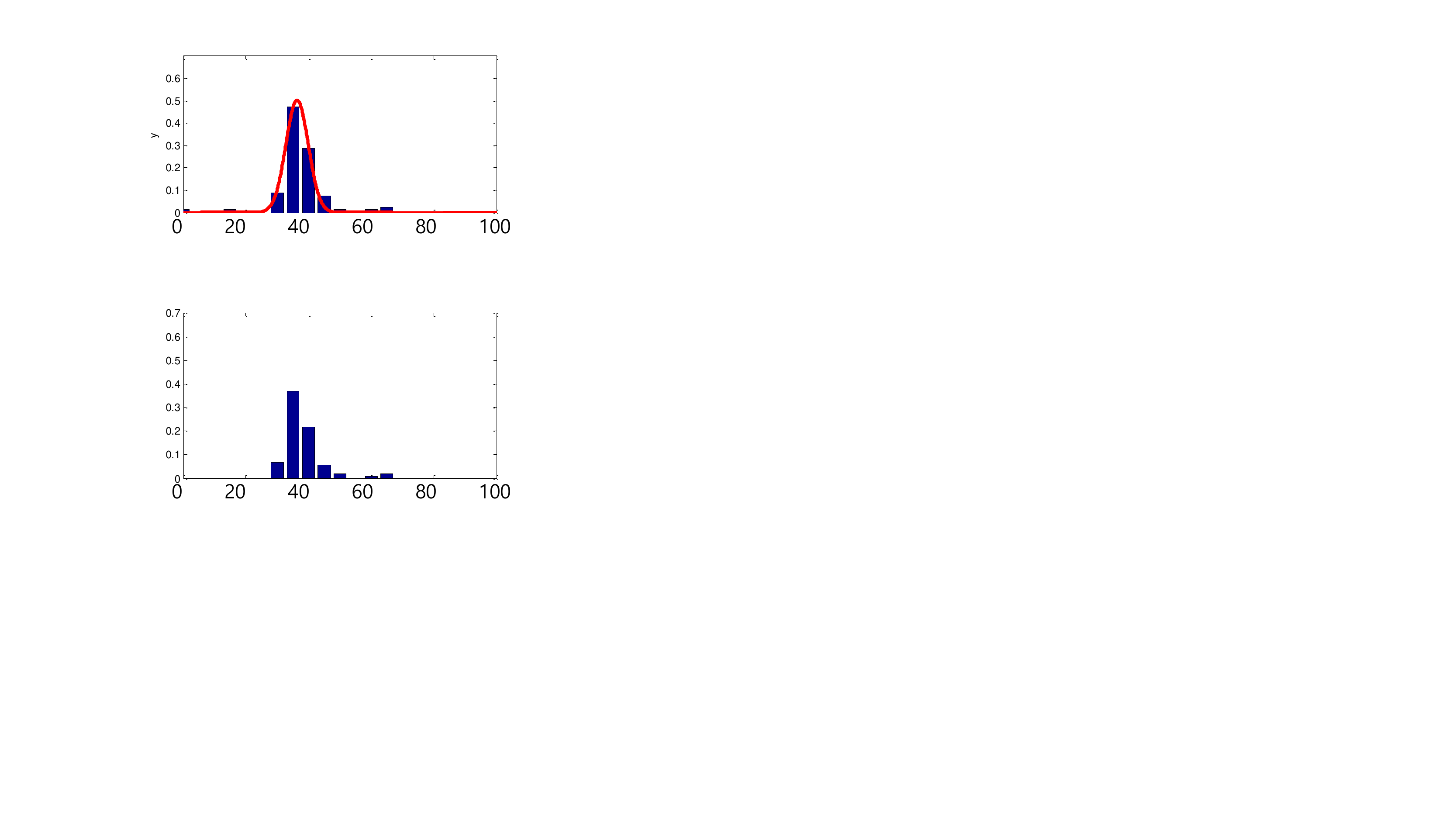}} 
		\vspace{-5pt}
		\caption{Comparison of inferred transition distributions and ground truth (DATA2 - Exit: \texttt{CAM1-ZONE2} , Entry: \texttt{CAM2-ZONE1}).}
		\label{fig_top_MCT}
	\end{figure*}   
				
	$\\$
	\noindent \textbf{Camera topology initialization results}
	
	After the CAM-to-CAM topology inference, we infer the topology of entry and exit zones.
	We found two valid links between \texttt{CAM1} and \texttt{CAM2} in both the datasets. 
	The first link is Exit (\texttt{CAM1-Zone2}) -- Entry (\texttt{CAM2-Zone1}) and the second link is Exit (\texttt{CAM2-Zone1}) -- Entry (\texttt{CAM1-Zone2}). Through this step, invalid pairs of cameras and zones are detected and ignored. In the next step, we iteratively update the valid links between zones and build a final camera topology map of the camera network.

	\begin{table}[t]
		      	\centering
		      	\setlength\tabcolsep{3.5pt}
		      	\caption{Rank-1 accuracies at different stages of camera topology initialization.}
		      	\label{tabel_1}
		      	\begin{tabular}{c|c|c|c}
		      		\noalign{\hrule height 1pt}
		      		& CAM-to-CAM & Zone-to-Zone & Iteration 1 \\ \hline
		      		DATA1 (\texttt{CAM1 - CAM2}) & 72.6\%     & 76.7\%       & 93.2\%      \\ \hline
		      		DATA2 (\texttt{CAM1 - CAM2}) & 69.8\%     & 70.8\%       & 91.5\%      \\ \noalign{\hrule height 1pt}
		      	\end{tabular}
	\end{table}

	After evaluating the Zone-to-Zone topology inference, we iteratively update the valid links between zones. 
	In the case of \texttt{NLPR MCT} dataset~\cite{Chen2015An}, valid Zone-to-Zone links converged within the second iteration.
	Table~\ref{tabel_1} shows the person re-identification accuracy according to each proposed topology initialization step.
	Our method shows a superior re-identification performances: 93.2\% and 91.5\%, for the camera pair \texttt{CAM1-CAM2}.
	
	We compared our method to several previous topology inference methods~\cite{makris2004bridging, niu2006recovering,chen2014object}.
	The previous methods showed unclear and noisy transition distributions as shown in Fig.~\ref{fig_top_MCT} (a-c).
	Even though the transition distribution inferred by Chen's method~\cite{chen2014object} shows a peak of the distribution, the overall distribution is too noisy to utilize.
	On the other hand, our result is very close to the ground truth and shows a very clear distribution as shown in Fig.~\ref{fig_top_MCT} (d,e).

   \section{Discussions}
   
   The proposed unified framework can infer camera network topology while it performs person re-identification.
   In this study, we assume that the transition distribution between the two links follows a single-mode Gaussian model since the model works well in most cases.
   In general, most previous study~\cite{makris2004bridging,niu2006recovering} also follow the same assumption.  
   Although it works well in most cases, it may fail in more complex situations, e.g., multiple paths between two zones. We will consider this issue in further studies.
   		
	\section{Conclusions}
				
	\label{sec:conclusion} 
	
	In this study, we proposed a unified framework to automatically solve both person re-identification and camera network topology inference problems. In addition, to validate the performance of person re-identification in the practical large-scale surveillance scenarios, we provided a new person re-identification dataset called~\texttt{SLP}. We qualitatively and quantitatively evaluated and compared the performance of the proposed framework with state-of-the-art methods using public and our own datasets. The results show that the proposed framework is promising for both person re-identification and camera topology inference and superior to other frameworks in terms of both speed and accuracy.

	\appendix
	\section{Computational Complexity\\ of the Proposed Method}

	\begin{figure}[t]
		\centering
		{\includegraphics[width=0.7\columnwidth]{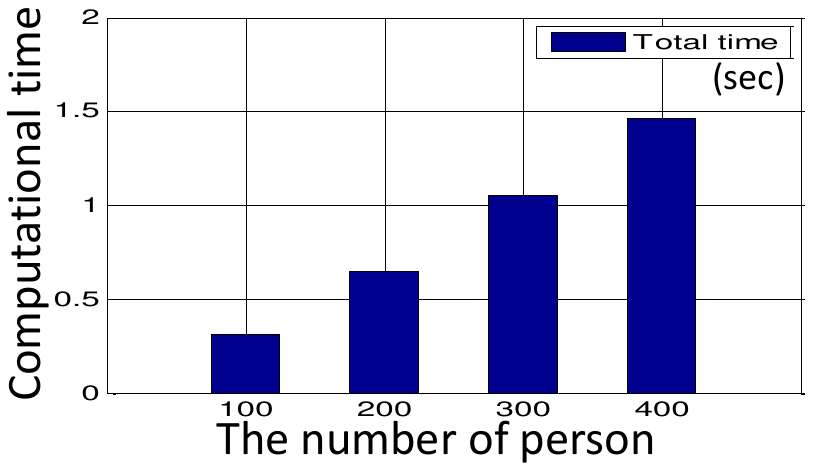}}
		\caption{Computation time of the proposed method based on the number of people in each camera
			We fixed the number of appearances of each person used for multi-shot matching $(K=30)$. The computation time includes random forest training and testing time.}
		\label{fig_complex}
	\end{figure}
				
	As we mentioned in the paper, the traditional exhaustive re-identification approaches examine every person pair between two different cameras $(c_{A}, c_{B})$. Assuming that the numbers of objects in $c_{A}$ and $c_{B}$ are the same as $N$ and each object contains the same number of appearances as $K$, it is obvious that the exhaustive approaches need $({N}^{2}{K}^{2})$ time for appearance comparisons for re-identification. Using the Big-O notation, the computational complexity of the exhaustive approach is defined as $O\left({N}^{2}{K}^{2}\right)$.
	
	On the other hand, in the proposed method, the tree structure of the random forest method makes the multi-shot person re-identification very fast.
	In return, the random forest containing multiple decision trees needs training time. 
	If we denote the number of decision trees and the random forest training time as $M$ and $T$ respectively, we need $\left(MNK\log N\right)$ time for the appearance comparisons and $T$ for forest training. 
	Here, note that the number of decision trees is a constant value. We fixed it as $10$ in our experiments. Therefore, we can ignore the number of the decision trees $M$ in the Big-O notation since $M \ll N,K$. In addition, the random forest training time in the training stage can be also be ignored because the training time $T$ \textit{linearly} increases when the number of cameras in a camera network increases whereas the test time \textit{exponentially} increases when the number of cameras increases. Actually, the computational complexity of the test stage is proportional to the number of possible camera pairs: ${ _{ L }{ C }_{ 2 } }$, where $L$ is the number of cameras in a camera network. 
	For that reason, the training time $T$ can be ignored when the camera network is large. 
	As a result, the complexity of proposed method is approximated as $O\left(NK\log N \right)$.
	
	Fig.~\ref{fig_complex} shows the computation time according to the number of people in each camera in the two-camera re-identification case. 
	As we can see, the computation time of the proposed method increases linearly, not exponentially, as the number of people increases.

	{\small
		\bibliographystyle{IEEEtran}
		\bibliography{TCSVT_bib}
	}

\begin{IEEEbiography}{Yeong-Jun Cho} received the B.S. degree in information and communication engineering from Korea Aerospace University and the M.S. degree in information and communications from Gwangju Institute of Science and Technology (GIST), in 2012 and 2014, respectively. He is currently pursuing the ph.D. degree as a member of the Computer Vision Laboratory in GIST. His research interests include computer vision topics such as person re-identification, multi-object tracking, object detection and medical image analysis.
\end{IEEEbiography}

\begin{IEEEbiography}{Su-A Kim} received the B.S. degree from the Department of Electronic and IT Media Engineering from Seoul National University of Science and Technology in 2013 and the M.S. degree in School of Information and Communications from Gwangju Institute of Science and Technology (GIST) in 2015. She is currently a scientific researcher at Intel Visual Computing Institute and pursuing the Ph.D. degree in Saarland University. Her research interests include key research topics in computer vision such as person re-identification, 3D object detection and pose estimation, and 4D light field segmentation. 
\end{IEEEbiography}

\begin{IEEEbiography}{Jae-Han Park} received the B.S. degree in electrical engineering and computer science from Gwangju Institute of Science and Technology (GIST) in 2015. He is currently pursuing M.S degree as a member of Computer Vision Laboratory in GIST. His research interests include person re-identification, video summarization and age estimation using deep learning.
\end{IEEEbiography}

\begin{IEEEbiography}{Kyuewang Lee} received the B.S. degree in Electrical Engineering and Computer Science from Gwangju Institute of Science and Technology (GIST) in 2016. He is currently pursuing the integrated M.S./Ph.D. degree as a member of the Perception and Intelligence Laboratory in Seoul National University. His research interests include key research topics in computer vision, such as visual object tracking, multi-object tracking, and person re-identification.
\end{IEEEbiography}

\begin{IEEEbiography}{Kuk-Jin Yoon}
	received the B.S., M.S., and Ph.D. degrees in Electrical Engineering and Computer Science from Korea Advanced Institute of Science and Technology (KAIST) in 1998, 2000, 2006, respectively. He was a post-doctoral fellow in the PERCEPTION team in INRIAGrenoble, France, for two years from 2006 and 2008 and joined the School of Information and Communications in Gwangju Institute of Science and Technology (GIST), Korea, as an assistant professor in 2008. He is currently an associate professor and a director of the Computer Vision Laboratory in GIST.
\end{IEEEbiography}

%




\end{document}